\documentclass[journal]{IEEEtran}
\usepackage{amssymb}
\usepackage{color}
\usepackage{cite}
\usepackage{booktabs}
\usepackage{multirow}
\usepackage{multicol}
\usepackage{amsfonts}
\usepackage{textcomp}
\usepackage{changepage}
\usepackage{algorithm} 
\usepackage{algorithmic}
\usepackage{float}

\usepackage{tabularx}
\usepackage{amsmath,amsfonts}
\usepackage{algorithmic}
\usepackage{algorithm}
\usepackage{array}
\usepackage{textcomp}
\usepackage{stfloats}
\usepackage{url}
\usepackage{verbatim}
\usepackage{cite}
\usepackage{graphicx}
\usepackage{subfigure}
\usepackage{makecell}
\usepackage[table]{xcolor}
\usepackage[normalem]{ulem}
\usepackage{array}
\useunder{\uline}{\ul}{}
\usepackage[
colorlinks=true,
citecolor=blue,
linkcolor=red,
urlcolor=pink
]{hyperref}
\definecolor{MyForestGreen}{rgb}{0.13, 0.55, 0.13} 
\begin{document}

\title{SARVLM: A Vision Language Foundation Model for Semantic Understanding in SAR Imagery}
\author{Qiwei~Ma, Xukun~Lu, Wang~Liu, 
Puhong~Duan, ~\IEEEmembership{Member,~IEEE}, 

Xudong~Kang, ~\IEEEmembership{Senior Member,~IEEE}, Shutao~Li, ~\IEEEmembership{Fellow,~IEEE}, 

\thanks{This work was supported in part by the National Natural Science Foundation of China under Grant 62525108 and Grant 62371185, in part by the National Key Research and Development Program of China under Grant 2021 YFA0715203, in part by the Science and Technology Inovation Program of Hunan Province under Grant 2024RC1030 and Grant 2023RC3124, in part by the Project of Yuelushan Center for Industrial Innovation under Grant 2025YCII0202. \textit{(Corresponding author: Puhong Duan.)}}

\thanks{Q. Ma, W. liu, P. Duan, and S. Li are with the School of Artificial Intelligence and Robotics, Hunan University, Changsha, 410082, China (e-mail: maqiwei@hnu.edu.cn; liuwa@hnu.edu.cn; puhong\_duan@hnu.edu.cn; shutao\_li@hnu.edu.cn)}

\thanks{X. Kang is with the School of Artificial Intelligence and Robotics, Hunan University, Changsha, 410082, China, and also with the Yuelushan Center for Industrial Innovation, Changsha, 410082, China. e-mail:(xudong\_kang@163.com)}

\thanks{X. Lu is with the School of Medical Information Engineering, Jining Medical University, Rizhao, Shandong Province, 276026, China (e-mail: luxukun@hnu.edu.cn)}
}

\markboth{Submitted to IEEE Transactions on Image Processing, ~Vol.~X, No.~X, 2026}
{Shell \MakeLowercase{\textit{et al.}}:TIP SARVLM}
%



\maketitle

\begin{abstract}
Synthetic Aperture Radar (SAR) is a critical imaging modality due to its all-weather operational capability. 
Although recent advances in self-supervised learning and masked image modeling (MIM) have enabled SAR foundation models, these approaches primarily focus on low-level visual features and often neglect multi-modal representation. Moreover, multimodal data for SAR is scarce, limiting the development of robust cross-modal models.
To address this limitation, we construct SARVLM-1M, a large-scale vision-language dataset comprising over one million image-text pairs aggregated from existing datasets. 
Furthermore, to mitigate the substantial differences between SAR and natural imagery, we propose a two-stage domain transfer training strategy that leverages optical remote sensing data as an intermediate bridge, facilitating effective knowledge transfer from natural images to SAR domains. 
Based on this strategy, we develop SARVLM, the first vision-language foundation model tailored for SAR, consisting of SARCLIP and SARCap. 
In addition, an ensemble strategy is utilized to improve the cross-scene generalization capability of the model. Moreover, SARDet and SARRot further validate the capability of the proposed framework in object detection. 
Extensive experiments on 13 benchmarks across image-text retrieval, target recognition, zero-shot classification, object detection, semantic localization, and image captioning demonstrate the superior feature extraction and interpretation capabilities of SARVLM. It consistently outperforms state-of-the-art vision-language models and advances semantic understanding in SAR imagery.
Code and datasets will be released on \url{https://github.com/KlayMa527/SARVLM.git}.
\end{abstract}


\section{Introduction}

SAR is a high-resolution imaging technology with all-weather, day-night operability and strong penetration capabilities, widely applied in military, environmental, maritime, and disaster monitoring tasks~\cite{flood}. Compared with optical imagery, SAR images are characterized by speckle noise, geometric distortions, and limited semantic textures, which pose significant challenges for downstream tasks such as object detection~\cite{Speckle_SSD,SSDD}, target recognition~\cite{ding_sar_atr,RABS,SARATR,PCM}, computational imaging~\cite{IALM}, change detection~\cite{zhang_SAR_change}, and image classification \cite{CCSL}. These challenges highlight the need for robust and generalizable feature representations specifically tailored to SAR data.
\begin{figure}[!t]
	\centering
	\includegraphics[width=0.9\linewidth,keepaspectratio]{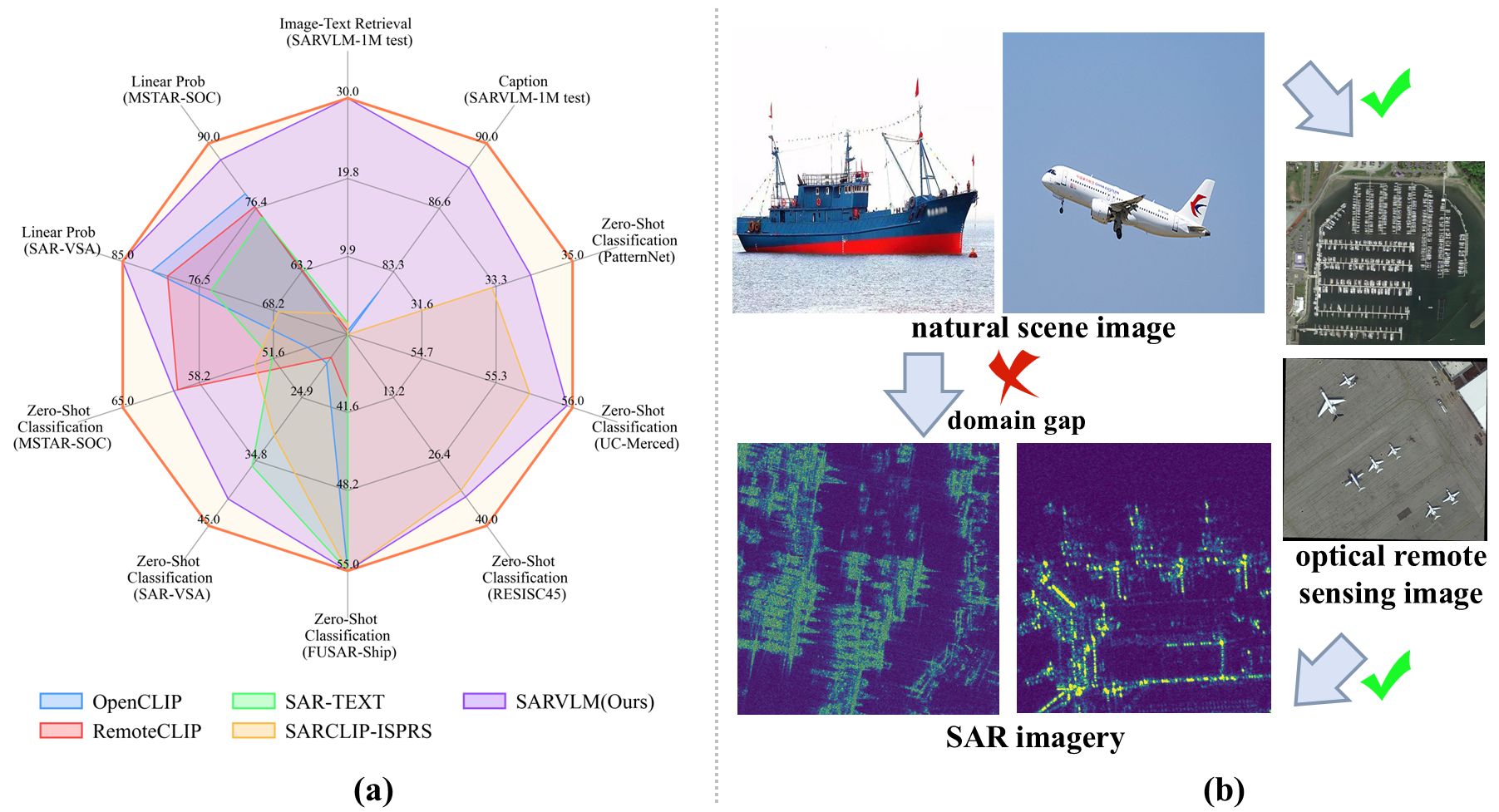}
	\caption{(a) Compared with vision-language foundation models pretrained on natural images, optical remote sensing imagery, or SAR imagery, the proposed SARVLM delivers the most competitive and well-balanced performance across diverse downstream tasks.
		(b) A key challenge of SAR understanding lies in the substantial domain gap between natural scene images and SAR imagery.
		Optical remote sensing imagery serves as an effective intermediate domain, as it shares higher-level scene semantics with natural images while being more closely related to SAR data in remote sensing scenarios.
		This motivates our two-stage progressive pre-training strategy for bridging the gap from natural imagery to SAR imagery.}
	\label{fig_teaser}
\end{figure}

\begin{figure*}[t!]
	\centering
	\includegraphics[width=\linewidth]{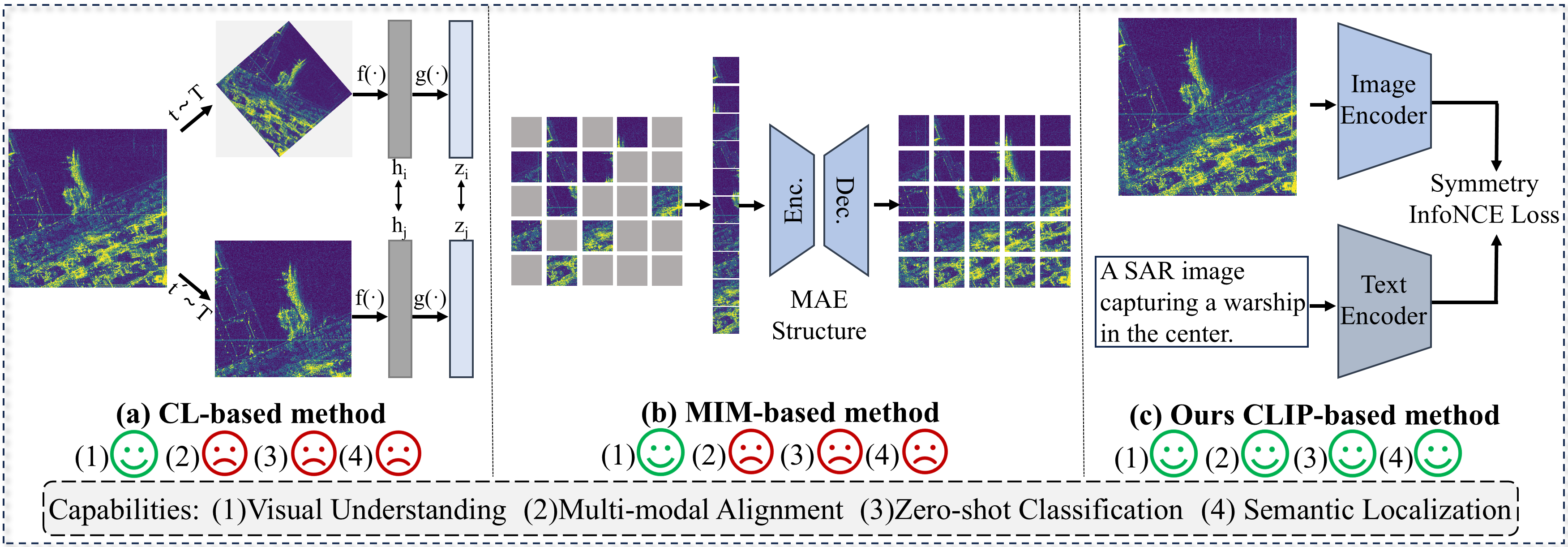}
	\caption{The paradigm for foundation model: (a) CL-based methods, (b) MIM-based methods, (c) CLIP-based methods.}
	\label{fig_paradigm}
\end{figure*}

Recent developments in vision foundation models (VFMs) have led to promising generalization capabilities across domains. As illustrated in Fig.~\ref{fig_paradigm}, existing VFMs can be broadly categorized into three paradigms: (a) contrastive learning (CL-based) methods such as SimCLR \cite{SimCLR}, which aim to learn discriminative representations by aligning augmented views of the same image; (b) masked image modeling (MIM-based) methods that reconstruct occluded image regions to learn spatial representations; and (c) CLIP-based approaches \cite{CLIP} that align images and texts via contrastive loss on large-scale paired data, showing remarkable transferability in cross-modal tasks.

The self-supervised training paradigm has recently driven the development of several vision foundation models in remote sensing, including RS-BYOL \cite{RS-BYOL}, ScaleMAE \cite{ScaleMAE}, and Cross-scale MAE \cite{Cross-Sclae-MAE}. Notably, ScaleMAE \cite{ScaleMAE} focuses on learning multi-scale representations by reconstructing both low- and high-frequency components across defined spatial scales. Similarly, Cross-scale MAE \cite{Cross-Sclae-MAE} enforces consistency across scales using a combination of contrastive and generative losses to enhance self-supervised remote sensing representations. However, these methods predominantly address the extraction of visual features. Inspired by CLIP \cite{CLIP}, RemoteCLIP \cite{RemoteCLIP} pioneered the first vision language foundation model (VLFM) in remote sensing, aiming to capture semantic features by aligning image-text pairs. Subsequent works like RS5M \cite{RS5M} and SkyScript \cite{Skyscript} have further explored this aspect. However, these models are pre-trained on optical image, leaving the SAR modality underexplored due to its distinct characteristics and lack of large-scale image-text datasets in SAR domain.

In recent years, the growing interest in SAR modality has led to the creation of various large-scale SAR datasets \cite{SARDet-100K, SARATR, FAIR-CSAR, ATRNet-STAR}. Specifically, SARDet-100K consolidates existing SAR datasets to form a large collection encompassing ships, aircraft, cars, bridges, tanks, and harbors. Building upon this, SARATR-X extends SARDet-100K by integrating additional classification datasets to create the larger SARDet-180K. Both SARDet-100K \cite{SARDet-100K} and SARATR-X \cite{SARATR} then employ MIM training strategy to establish their respective foundation models. Furthermore, recent advanced methods have to address the application of multi-modal LLMs in the SAR field. Particularly, SARChat-2M \cite{SARChat-Bench-2M} introduces a substantial benchmark with two million multimodal dialogues, facilitating intelligent interpretation of SAR imagery through LLM-based conversational paradigms. A comparable initiative has been undertaken with SARLANG-1M \cite{SARLANG-1M}. However, despite their impressive performance, existing SAR vision foundation models remain limited by the lack of textual annotations in current datasets, hindering their ability to capture rich semantic information fully. In addition, although SARCLIP-ISPRS~\cite{SARCLIP-ISPRS} has demonstrated promising results, it overlooks the substantial domain gap between natural optical and SAR imagery. As a result, the transfer of visual-language priors from the optical domain to the SAR domain remains insufficient, which consequently limits its performance.

To bridge this gap, as shown in Fig.~\ref{fig_teaser}, we propose \textbf{SARVLM} consisting of SARCLIP and SARCap,  a vision-language foundation model designed specifically for SAR imagery. Then, we construct \textbf{SARVLM-1M}, a large-scale SAR image-text dataset comprising 1.7 million pairs across diverse object categories and land cover types. These pairs are generated by leveraging domain knowledge, spatial rules, and LLM-based text synthesis strategies. Built upon the SARVLM-1M dataset, a two-stage progressive domain transfer training strategy is developed to transfer knowledge from the optical domain to the SAR domain, thereby facilitating semantic alignment between SAR imagery and language. As shown in Fig.~\ref{fig_framework}, this approach enables the extraction of comprehensive general features from SAR images and significantly enhances their semantic understanding, thereby elevating performance for SAR interpretation.

The key contributions are summarized as follows:
\begin{itemize}
	\item We construct SARVLM-1M, a large-scale vision-language dataset comprising 1.7 million image-text pairs that cover diverse objects and land cover types.
	\item We propose a two-stage domain transfer training strategy, enabling effective alignment between SAR images and textual data while improving cross-modal representation.
	\item We propose SARVLM, a vision-language foundation model specifically designed for SAR imagery, enabling SAR-specific cross-modal alignment, target recognition, zero-shot classification, object detection, semantic localization, and image captioning.
	\item Extensive experiments demonstrate that SARVLM consistently surpasses existing state-of-the-art VLFMs on multiple downstream tasks, highlighting its strong generalization and semantic representation capabilities in the SAR domain.
\end{itemize}

\section{Related Work}
\subsection{Vision Language Model for Remote Sensing}
Vision-language models have significantly advanced remote sensing through the development of multi-modal techniques. These methods broadly fall into two categories: contrastive methods and generative methods. Among contrastive approaches, CLIP \cite{CLIP} stands out as a pioneering work, employing a two-tower architecture to align visual and language features through extensive data from Internet. Inspired by CLIP, several studies have adapted this paradigm to remote sensing, yielding models such as RemoteCLIP \cite{RemoteCLIP}, GeoRSCLIP \cite{RS5M}, SkyScript \cite{Skyscript}, RSMCLIP \cite{rethinking} and Mall et al. \cite{mall2024remote}. Notably, RemoteCLIP \cite{RemoteCLIP} enhances pre-training by converting detection and segmentation annotations into image captions, facilitating CLIP-style contrastive learning for image-text alignment in remote sensing. In the realm of generative methods, works like GeoChat \cite{Geochat}, EarthGPT \cite{EarthGPT}, and LHRS-Bot \cite{LHRS-Bot} implement auto-regressive large language model (LLM) architectures for vision-text alignment. For instance, GeoChat \cite{Geochat} introduces a multimodal model with LLaVa architecture in remote sensing, achieving multi-granularity alignment through parameter-efficient fine-tuning. However, most of these existing methods primarily focus on the optical modality in remote sensing, largely overlooking investigations in the SAR domain.
\subsection{SAR Foundation Model}

Foundation models pretrained on large-scale data can capture generalizable visual representations and effectively support a wide range of downstream tasks~\cite{li2026graph}. In remote sensing, recent advancements have seen the utilization of self-supervised learning and MIM techniques, as demonstrated by approaches such as RingMo \cite{RingMo}, SatMAE \cite{SatMAE}, ScaleMAE \cite{ScaleMAE}, and Cross-scale MAE \cite{Cross-Sclae-MAE}. These models have been widely applied in tasks like aerial object detection and target recognition.
\begin{figure*}[t!]
	\centering
	\includegraphics[width=\linewidth]{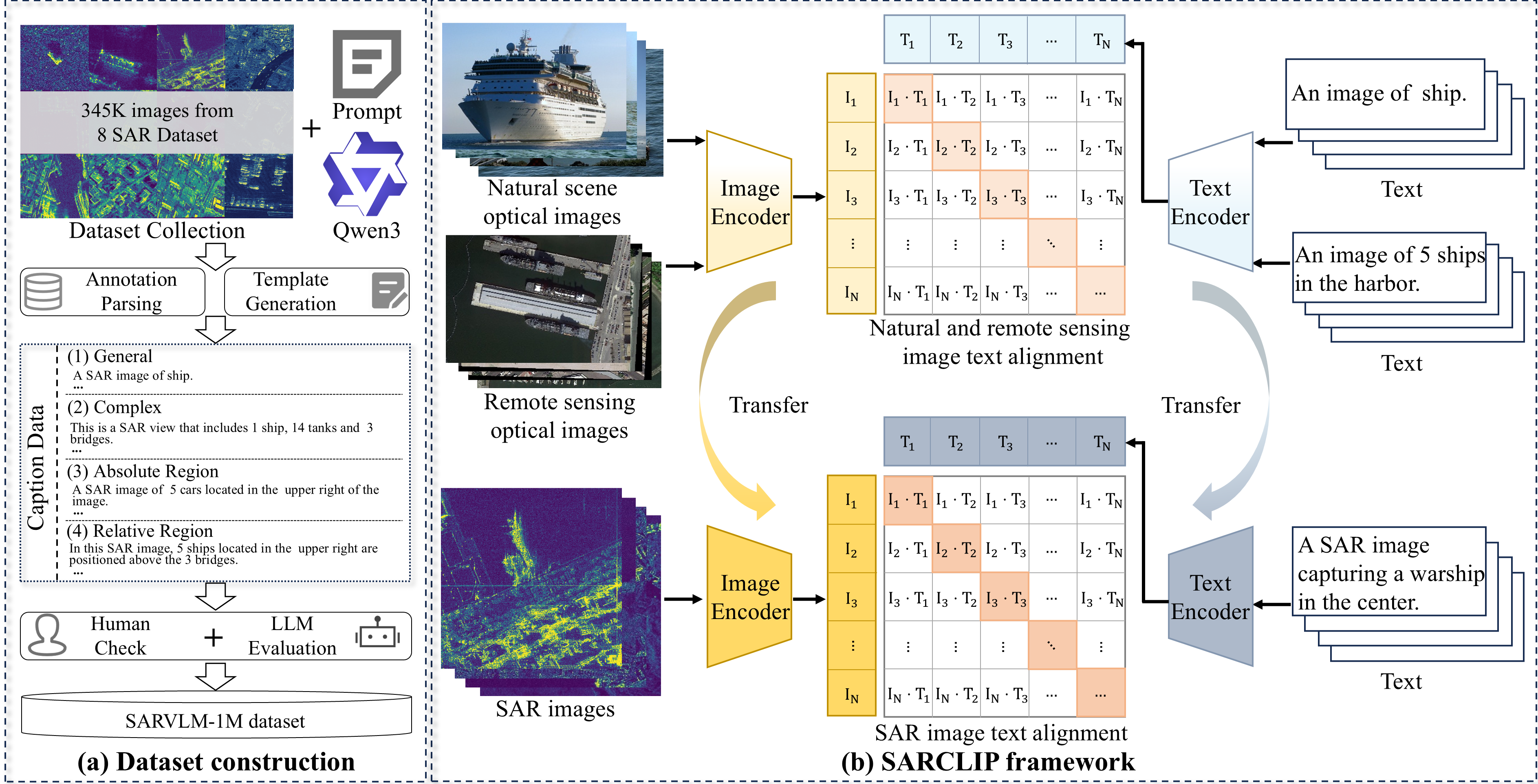}
	\caption{(a) Illustration of workflow for SARVLM-1M dataset construction; (b) Two-stage domain transfer training strategy for SARCLIP.}
	\label{fig_framework}
\end{figure*}

With the increasing availability of SAR imagery, a multitude of datasets have emerged, including SARDet-100K \cite{SARDet-100K}, SARATR-X \cite{SARATR}, SAR-JEPA \cite{SAR-JEPA}, FAIR-CSAR \cite{FAIR-CSAR}, and ATRNet-STAR \cite{ATRNet-STAR}. Specifically, SARDet-100K constructs SAR foundation model through MIM training, initially pretraining on aerial view images before transferring to SAR imagery. Building upon this, SARATR-X leverages various classification datasets to build the SARDet-180K dataset, subsequently yielding a MIM-based SAR foundation model. SARMAE~\cite{sun_sarmae} proposes foundation model based on complex-valued, introducing physical interpretability through polarimetric decomposition. Furthermore, SARLANG-1M \cite{SARLANG-1M} utilizes LLMs for SAR image interpretation, while SARChat \cite{SARChat-Bench-2M} supports key tasks such as visual understanding and object detection in SAR imagery. Nevertheless, a common limitation among these existing SAR foundation models is their primary focus on low-level image features, often failing to capture deeper semantic information and realize multi-modal alignment within SAR images.

\section{Methodology}
This section introduce the paradigm of our framework, dataset construction approach and training strategy for SARCLIP and SARCap.
\subsection{Problem Definition}

In this section, we investigate the paradigm of learning joint representations from SAR images and their corresponding textual descriptions. Specifically, we construct the SARVLM-1M dataset
$\mathcal{D}=\{(\mathbf{I}_{i}, \mathbf{T}_{i})\}_{i=1}^{M}$, consisting of SAR images $\mathbf{I}_i\in\mathcal{R}^{H \times W}$ with corresponding textual descriptions $\mathbf{T}_{i} \in \mathcal{T}$. 
As shown in Fig.~\ref{fig_framework}, our objective is to learn a pair of modality-specific encoders that project both SAR images and text into a shared semantic space. 
Specifically, we define a visual encoder $f_{v}:\mathcal{R}^{H \times W} \rightarrow\mathcal{R}^{d}$ that maps the input SAR image to a $d$-dimensional visual feature embedding $\mathbf{z}_{v}^{i}=f_{v}({\mathbf{I}_{i}})$, 
and a textual encoder $f_{t}:\mathcal{R}^{\mathcal{T}} \rightarrow\mathcal{R}^{d}$ that maps the textual input to a corresponding textual embedding $\mathbf{z}_{t}^{i}=f_{t}({\mathbf{T}_{i}})$. 
The goal is to align the embeddings $\mathbf{z}_{v}^{i}$ and $\mathbf{z}_{t}^{i}$ of matched image-text pairs in a common representation space, such that semantically similar inputs across modalities are embedded close to each other.

\begin{figure}[t!]
	\centering
	\includegraphics[width=0.9\linewidth]{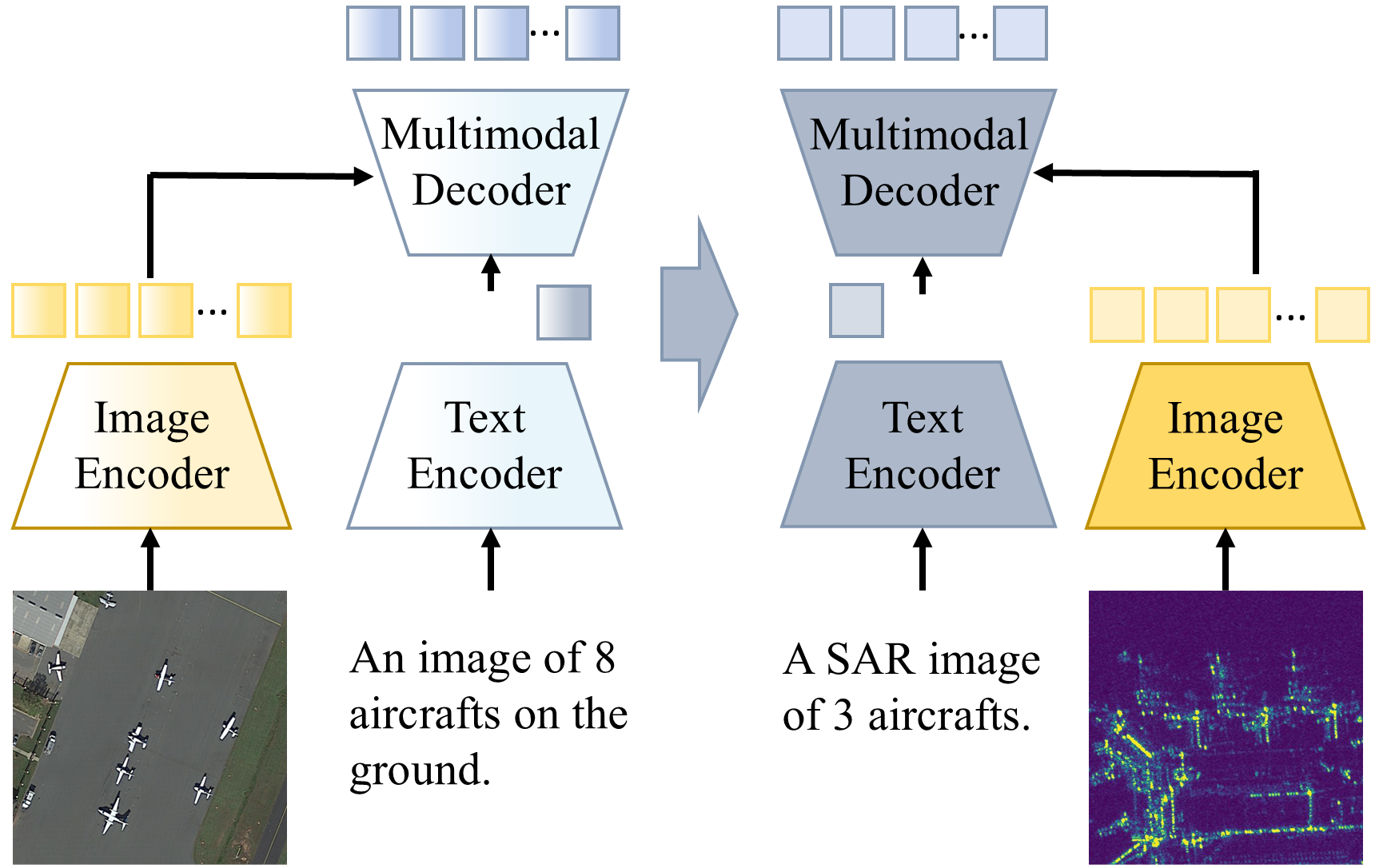}
	\caption{The framework of SARCap method.}
	\label{fig_caption_framework}
\end{figure}

In addition to joint representation learning, as shown in Fig.~\ref{fig_caption_framework}, SARVLM-1M supports the Caption task, where the goal is to generate a textual description $\hat{\mathbf{T}}_{i}$ for a given SAR image $\mathbf{I}_i$. 
Formally, this can be expressed as learning a function $g: \mathcal{R}^{H \times W} \rightarrow \mathcal{T}$ such that $\hat{\mathbf{T}}_i = g(\mathbf{I}_i)$.

Therefore, these formulations enable the model to bridge the modality gap between SAR images and natural language, thereby supporting a wide range of downstream tasks in the SAR domain, 
including cross-modal retrieval, target recognition, zero-shot classification, semantic localization, and image captioning.
\begin{figure}[t!]
	\centering
	\includegraphics[width=0.9\linewidth,keepaspectratio]{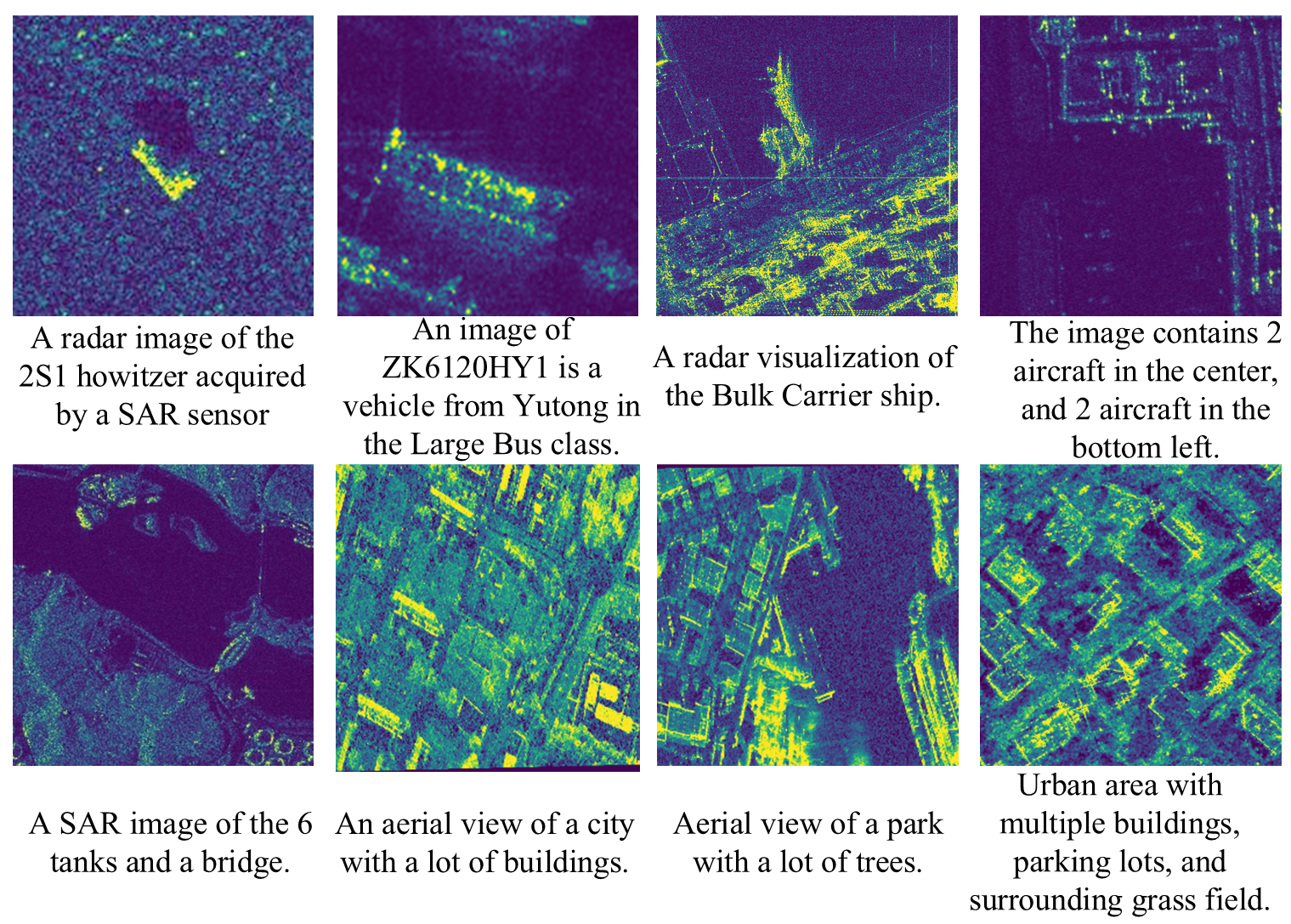}
	\caption{Image-text pair examples of the SARVLM-1M dataset.}
	\label{fig_dataset_demo}
\end{figure}

\subsection{SARVLM-1M Dataset Construction}
To address the challenges in building large-scale SAR vision-language datasets, we construct SARVLM-1M by aggregating existing classification, detection, and caption datasets in the SAR domain. As summarized in Table~\ref{table:datasets}, several large-scale SAR datasets have been released in the past two years. To enhance the diversity of textual descriptions and avoid overly repetitive captions, we design a multi-level template system covering general, complex, absolute region, and relative region descriptions. Moreover, these templates are enriched using large language model Qwen3~\cite{yang2025qwen3} to generate diverse linguistic variations and improve the expressiveness of the captions. Additionally, the samples of image-text pairs from SARVLM-1M dataset are shown in Fig.~\ref{fig_dataset_demo}, and the workflow is introduced as follows:

\textbf{(1) General descriptions.} We employ simple templates such as \textit{"A SAR image of the [class]"}, where \textit{[class]} is replaced with category names from classification datasets or object types and counts from detection datasets. These templates provide straightforward, unambiguous descriptions of image contents.

\textbf{(2) Complex descriptions.} To enrich the linguistic diversity, we utilize more elaborate templates such as \textit{"A SAR image reveals the distinct texture and structure of the [class]."} These templates capture fine-grained semantic details and enhance model robustness to varied textual expressions.

\textbf{(3) Absolute region descriptions.} Each image is partitioned into five regions: upper left, upper right, bottom left, bottom right, and center. We determine the target location by calculating the intersection-over-union (IoU) between annotated bounding boxes and these regions. Captions are generated using templates such as \textit{"A SAR image of [classes] located in the [location] of the image."}, explicitly encoding spatial information.

\textbf{(4) Relative region descriptions.} Spatial relationships between targets are described using relative templates such as \textit{"In this SAR image, the [class1] in the [location1] are positioned [relative\_direction] the [class2] in the [location2]."}, where \textit{[relative\_direction]} includes above, below, left, and right. This captures relational context and further enriches the semantic diversity.

\begin{table*}[t!]
	\setlength{\tabcolsep}{0.2mm}
	\renewcommand{\arraystretch}{1}
	\centering
	\caption{Overview of SARVLM-1M and its constituent datasets. Cls: Classification. Det.: Detection. Cap.: Caption. \# Train Imgs.: Number of training images. \# Train Caps.: Number of training captions. \# Val Imgs.: Number of validation images. \# Val Caps.: Number of validation captions. \# Test Pairs: Number of image-text pairs in testing set.}
	\begin{tabular*}{\textwidth}{@{\extracolsep{\fill}}
			>{\raggedright\arraybackslash}m{2.7cm}
			>{\centering\arraybackslash}m{0.8cm}
			>{\centering\arraybackslash}m{0.9cm}
			>{\centering\arraybackslash}m{0.9cm}
			>{\centering\arraybackslash}m{0.9cm}
			>{\centering\arraybackslash}m{0.9cm}
			>{\centering\arraybackslash}m{0.9cm}
			>{\centering\arraybackslash}m{0.9cm}
			>{\raggedright\arraybackslash}m{7cm}
		}
		\hline
		Dataset & Year & Task 
		& \makecell[c]{\# Train \\Imgs.} 
		& \makecell[c]{\# Train \\Caps.} 
		& \makecell[c]{\# Val \\Imgs.} 
		& \makecell[c]{\# Val \\Caps.} 
		& \makecell[c]{\# Test \\Pairs} 
		& Description \\
		\hline
		MSTAR \cite{MSTAR} & 1995 & Cls. & 3,046 & 15,230 & 9,855 & 49,275 & 180 & Contains X-band SAR imagery of military vehicles. \\
		SARSim \cite{SARSim_1,SARSim_2} & 2017 & Cls. & 21,168 & 105,840 & -- & -- & -- & Simulation dataset providing vehicle samples across 7 categories. \\
		OpenSARShip \cite{OpenSARship} & 2017 & Cls. & 26,679 & 133,395 & -- & -- & -- & Ship slices from European C-band Sentinel-1 satellite data. \\
		SAMPLE \cite{SAMPLE} & 2019 & Cls. & 5,380 & 26,900 & -- & -- & -- & Public X-band SAR dataset of 10 vehicle classes, with synthetic and real image pairs. \\
		ATRNet-STAR \cite{ATRNet-STAR} & 2025 & Cls.\&Det. & 68,091 & 340,455 & 29,284 & 146,420 & 6,667 & Large-scale SAR dataset offering 40 fine-grained vehicle target classes. \\
		SARDet-100K \cite{SARDet-100K} & 2024 & Det. & 94,493 & 472,465 & 10,492 & 52,460 & 2,783 & Compiled from 10 existing SAR detection datasets, encompassing 5 object classes. \\
		FAIR-CSAR \cite{FAIR-CSAR} & 2024 & Det. & 51,948 & 259,740 & 11,790 & 58,950 & 7,096 & Large-scale, fine-grained SLC SAR dataset covering 22 subcategories. \\
		SARLANG-1M Captions \cite{SARLANG-1M} & 2025 & Cap. & 9,191 & 31,968 & 3,939 & 13,682 & 3,902 & Over 45,000 SAR image captions based on SpaceNet6, DFC2023, and OpenEarthMap. \\ \hline
		\textbf{SARVLM-1M (ours)} & 2025 & Cap. & \textbf{279,996} & \textbf{1,385,993} & \textbf{65,360} & \textbf{320,787} & \textbf{20,628} & Comprises over 1.7 million image-text pairs, including ship, vehicle, aircraft, bridge, and other land covers. \\
		\hline
	\end{tabular*}
	\label{table:datasets}
\end{table*}

\begin{figure}[t!]
	\centering
	\includegraphics[width=0.9\linewidth, keepaspectratio]{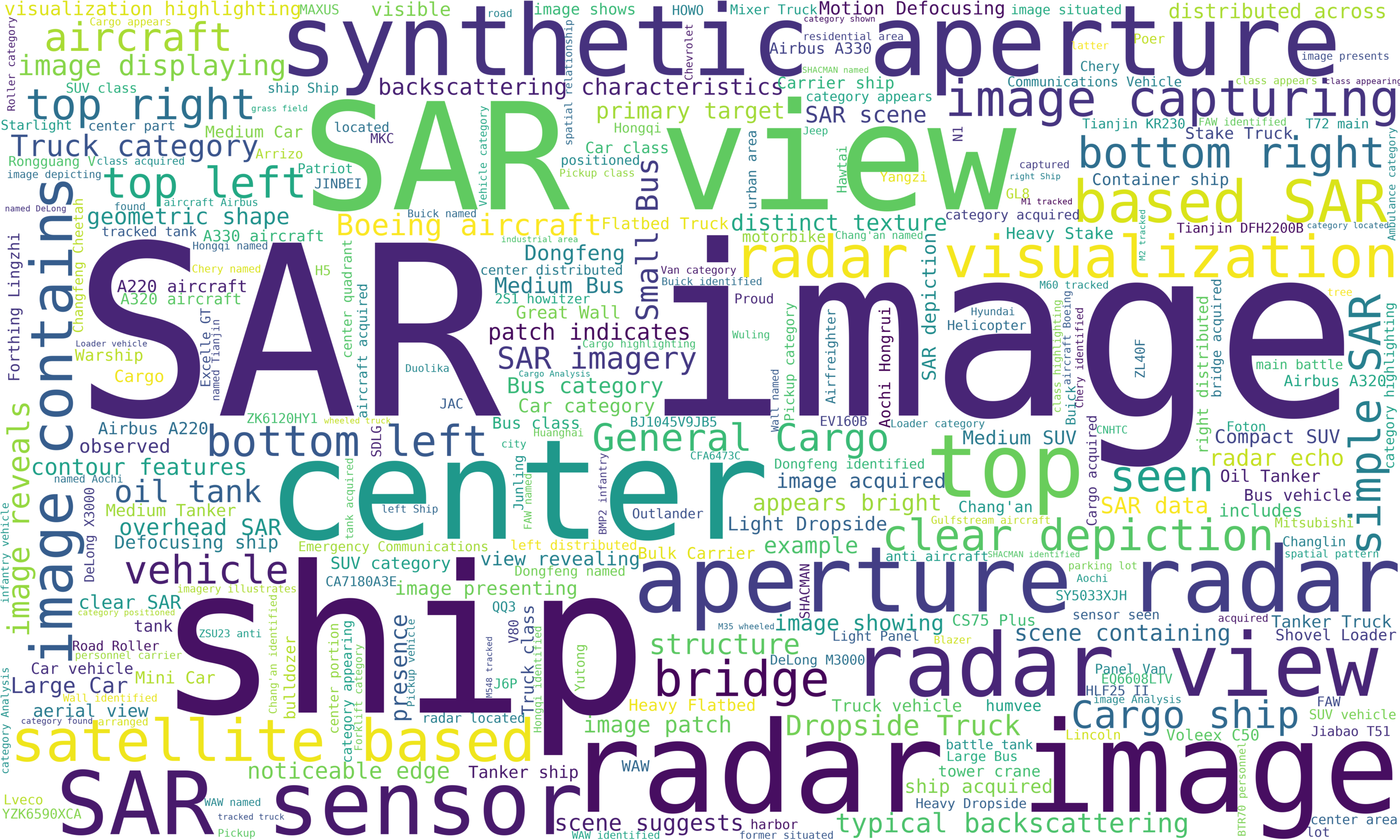}
	\caption{The word cloud of SARVLM-1M dataset.}
	\label{fig_word_cloud}
\end{figure}

\begin{figure}[t!]
	\centering
	\includegraphics[width=0.9\linewidth,keepaspectratio]{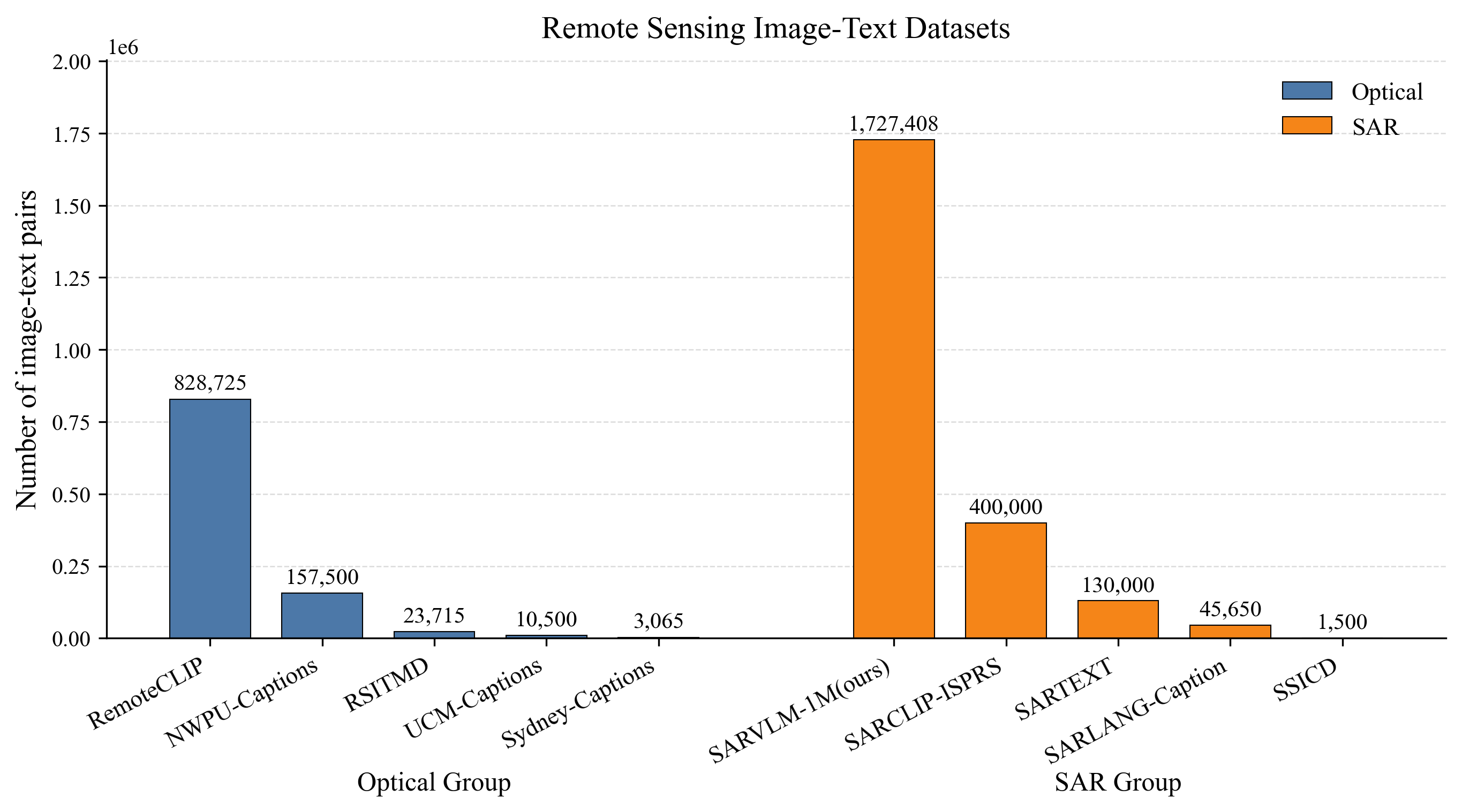}
	\caption{Number of image-text pairs in representative remote sensing datasets, grouped by modality (Optical vs. SAR).}
	\label{fig_dataset_pairs_comparison}
\end{figure}

By combining these multi-level templates, each SAR image is assigned multiple diverse captions. To ensure linguistic quality, we leverage a large language model to verify fluency and grammatical correctness. The resulting SARVLM-1M dataset comprises 279,996 images and 1,385,993 captions in the training set, and 20,628 image-text pairs in the test set. As illustrated in Fig.~\ref{fig_word_cloud}, the dataset covers a wide range of target types, including ships, vehicles, aircraft, bridges, and other land covers, providing rich and diverse multimodal supervision for SAR vision-language modeling. Moreover, as illustrated in Fig.~\ref{fig_dataset_pairs_comparison}, the number of image-text pairs in SARVLM is substantially larger than that in existing SAR-focused benchmarks, providing richer supervision for large-scale cross-modal representation learning in SAR field.

\subsection{Two-stage Domain Transfer Training.}
To effectively transfer knowledge from optical remote sensing imagery to SAR domains, we adopt a progressive two-stage fine-tuning strategy for both SARCLIP and SARCap. 
In the first stage, the visual encoder is initialized with weights pretrained on natural optical remote sensing datasets (e.g., RemoteCLIP~\cite{RemoteCLIP} with LoveDA, DOTA, and RSCID). 
This stage allows the model to capture general remote sensing visual representations, particularly for small-scale and densely distributed objects.
Formally, let $f_v^{(0)}$ denote the pretrained visual encoder on optical imagery. 
The first-stage fine-tuning updates the encoder parameters $\theta_v^{(1)}$ by minimizing the CLIP contrastive loss $\mathcal{L}_{\text{clip}}$:
\begin{equation}
	\mathbf{z}_v^i = f_v^{(0)}(\mathbf{I}_i^{\text{opt}})
\end{equation}
\begin{align}
	\mathcal{L}_{\text{clip}} = -\frac{1}{N} \sum_{i=1}^{N} \biggl\{\bigg[
	& \log \frac{\exp(\text{sim}(\mathbf{z}_v^i, \mathbf{z}_t^i)/\tau)}{\sum_{j=1}^{N} \exp(\text{sim}(\mathbf{z}_v^i, \mathbf{z}_t^j)/\tau)} \notag \\
	+ & \log \frac{\exp(\text{sim}(\mathbf{z}_t^i, \mathbf{z}_v^i)/\tau)}{\sum_{j=1}^{N} \exp(\text{sim}(\mathbf{z}_t^i, \mathbf{z}_v^j)/\tau)}
	\bigg] / 2\biggr\}
\end{align}
\begin{equation}
	\theta_v^{(1)} = \arg\min_{\theta_v} \frac{1}{N} \sum_{i=1}^{N} \mathcal{L}_{\text{clip}}(f_v^{(0)}(\mathbf{I}_i^{\text{opt}}), \mathbf{z}_t^i),
\end{equation}
where $\text{sim}(a,b)=\frac{a^\top b}{\|a\|\|b\|}$ is cosine similarity, $\tau$ denotes a temperature, $\mathbf{I}_i^{\text{opt}}$ denotes optical remote sensing images, $\mathbf{z}_t^i$ are their textual embeddings, and $N$ is the batch size. 
This step produces a visual encoder that already encodes general remote sensing semantics.

In the second stage, we progressively adapt the model to SAR imagery by fine-tuning on SARVLM-1M. 
The visual encoder parameters $\theta_v^{(2)}$ are initialized from the first stage $\theta_v^{(1)}$, and the textual encoder $f_t$ is trained jointly to align SAR images with text:
\begin{equation}
	\theta_v^{(2)}, \theta_t = \arg\min_{\theta_v, \theta_t} \frac{1}{M} \sum_{i=1}^{M} \mathcal{L}_{\text{clip}}(f_v(\mathbf{I}_i^{\text{SAR}};\theta_v), f_t(\mathbf{T}_i;\theta_t))
\end{equation}
where $\mathbf{I}_i^{\text{SAR}}$ and $\mathbf{T}_i$ are SAR images and their corresponding textual descriptions from SARVLM-1M, and $M$ is the number of SAR image-text pairs. 
This progressive adaptation enables effective transfer of semantic knowledge from optical to SAR imagery while preserving modality-specific feature representations.

For SAR image captioning, we adopt a similar two-stage approach. Let $g^{(0)}$ denote the pretrained captioning decoder from optical imagery. 
During SAR fine-tuning, we initialize the captioning decoder with $g^{(0)}$ and optimize its parameters $\theta_g$ using the captioning loss $\mathcal{L}_{\text{caption}}$:
\begin{equation}
	\theta_g = \arg\min_{\theta_g} \frac{1}{M} \sum_{i=1}^{M} \mathcal{L}_{\text{caption}}(g(\mathbf{I}_i^{\text{SAR}};\theta_g), \mathbf{T}_i)
\end{equation}
where 
\begin{equation}
	\mathcal{L}_{\text{caption}} = - \sum_{t=1}^{T} \log P(w_t | w_1, ..., w_{t-1}, z_v)
\end{equation}
$T$ is the caption length, $w_t$ is the $t$-th token, $z_v = f_v(\mathbf{I}_i^{\text{SAR}};\theta_v^{(2)})$ is the visual feature from SARCLIP, and $P(\cdot)$ denotes the predicted probability of $w_t$ conditioned on previous tokens and the visual feature. 
By initializing from $g^{(0)}$, the decoder leverages prior knowledge from optical imagery, while being adapted to SAR-specific visual patterns.

Overall, this progressive two-stage fine-tuning strategy provides a principled framework for cross-modal and cross-domain knowledge transfer. 
It enables SARCLIP and SARCap to bridge the modality gap between SAR images and natural language, supporting a wide range of downstream tasks, including cross-modal retrieval, target recognition, zero-shot classification, semantic localization, and image captioning.

\subsection{Ensemble Strategy}
To further incorporate the optical remote sensing prior into the SAR-domain model, a parameter-level ensemble strategy is adopted. Instead of performing score-level fusion during inference, the model weights obtained from the first-stage optical transfer and the second-stage SAR adaptation are directly fused in the parameter space. The corresponding model weights are denoted by
$\mathbf{\Theta}^{\mathrm{RS}}=\{\theta_k^{\mathrm{RS}}\}_{k\in\mathcal{I}^{\mathrm{RS}}}$ and
$\mathbf{\Theta}^{\mathrm{SAR}}=\{\theta_k^{\mathrm{SAR}}\}_{k\in\mathcal{I}^{\mathrm{SAR}}}$,
where $\mathcal{I}^{\mathrm{RS}}$ and $\mathcal{I}^{\mathrm{SAR}}$ denote the parameter index sets of the optical remote sensing model and the SAR-domain model, respectively. The ensemble ratio is defined as $\alpha \in [0,1]$, where $\alpha$ controls the contribution of the optical remote sensing model weights and $1-\alpha$ controls the contribution of the SAR-domain model weights.

A valid interpolation index set is defined as
\begin{equation}
	\mathcal{K}=
	\left\{
	k \,\middle|\,
	k\in \mathcal{I}^{\mathrm{RS}} \cap \mathcal{I}^{\mathrm{SAR}},
	\ \theta_k^{\mathrm{RS}},\theta_k^{\mathrm{SAR}} \in \mathbb{F}
	\right\},
\end{equation}
where $\mathbb{F}$ denotes the set of floating-point tensor weights, and $\theta_k^{\mathrm{RS}}$ and $\theta_k^{\mathrm{SAR}}$ are required to have identical shapes. Accordingly, only those parameters that are shared by the two models, have identical shapes, and are represented in floating-point format are blended. For each parameter index $k\in\mathcal{K}$, the fused weight is computed as
\begin{equation}
	\theta_k^{\mathrm{ens}}=\alpha \theta_k^{\mathrm{RS}}+(1-\alpha)\theta_k^{\mathrm{SAR}}.
\end{equation}
For the remaining parameters in the optical remote sensing model that do not satisfy the interpolation condition, the corresponding optical weights are retained:
\begin{equation}
	\theta_k^{\mathrm{ens}}=\theta_k^{\mathrm{RS}}, \quad k\in \mathcal{I}^{\mathrm{RS}}\setminus\mathcal{K}.
\end{equation}
The final fused model weights are denoted by $	\mathbf{\Theta}^{\mathrm{ens}}=\{\theta_k^{\mathrm{ens}}\}_{k\in\mathcal{I}^{\mathrm{RS}}}$. Based on the fused model weights $\mathbf{\Theta}^{\mathrm{ens}}$, zero-shot classification is performed over a class prompt set $\{T_c\}_{c=1}^{C}$:
\begin{equation}
	s_c=s(I,T_c), \quad
	\hat{y}=\arg\max_{c\in\{1,\dots,C\}} s_c,
\end{equation}
where $s_c$ denotes the matching score between the input image $I$ and the $c$-th class prompt $T_c$, and $\hat{y}$ denotes the predicted category.

In this way, the relative contributions of the optical-domain model and the SAR-adapted model are controlled by $\alpha$, while a single fused model is maintained for downstream inference. As a result, both the general scene semantics learned from optical remote sensing imagery and the SAR-specific scattering characteristics acquired during domain adaptation are preserved. The effect of the ensemble ratio $\alpha$ is further analyzed in Section~\ref{Sec_ensemble_ratio}.
\begin{table*}[t!]
	\centering
	\renewcommand\arraystretch{1.2}
	\setlength{\tabcolsep}{0.8mm}
	\caption{Retrieval performance on the SARVLM-1M test set and improvements over the SARCLIP baseline(\%). LAION~\cite{OpenCLIP} is an optical image dataset in the natural scene domain, while RS5M, Skyscript, and R3+D10+S4 are optical image datasets in the remote sensing domain. $^{\dagger}$, $^{\S}$, and $^{\ddagger}$ denote models pretrained on RS5M~\cite{RS5M}, Skyscript~\cite{Skyscript}, and R3+D10+S4~\cite{RemoteCLIP}, respectively.}
	\begin{tabular*}{\hsize}{@{\extracolsep{\fill}}lcllccccccc}
		\hline
		\multirow{2}{*}{Method} & \multirow{2}{*}{\makecell{Image\\Backbone}}   & \multirow{2}{*}{\makecell{Pretrain\\Data}}                & \multirow{2}{*}{\makecell{Tune\\On}}  & \multicolumn{3}{c}{Image to Text} & \multicolumn{3}{c}{Text to Image} & \multirow{2}{*}{\makecell{Mean\\Recall}}  \\ 
		\cmidrule(lr){5-7}\cmidrule(lr){8-10}
		&  &    &       & R@1       & R@5       & R@10      & R@1       & R@5      & R@10    &   \\ \hline
		OpenCLIP     & ResNet-50      & LAION       & --          & 0.04      & 0.18      & 0.38      & 0.06      & 0.20      & 0.38   & 0.21   \\
		OpenCLIP     & ViT-B-32       & LAION       & --          & 0.08      & 0.20      & 0.40      & 0.08      & 0.34      & 0.50   & 0.27   \\
		OpenCLIP     & ViT-L-14       & LAION       & --          & 0.14      & 0.34      & 0.70      & 0.10      & 0.56      & 1.12   & 0.49 \\
		GeoRSCLIP    & ViT-B-32       & LAION       & RS5M         & 0.06      & 0.30      & 0.62      & 0.12      & 0.54      & 0.96   & 0.43 \\
		GeoRSCLIP    & ViT-L-14       & LAION       & RS5M           & 0.06      & 0.24      & 0.50      & 0.20      & 0.74      & 1.36   & 0.52   \\
		GeoRSCLIP    & ViT-H-14       & LAION       & RS5M          & 0.06      & 0.22      & 0.48      & 0.20      & 0.94      & 1.32   & 0.54   \\
		SkyCLIP      & ViT-B-32       & LAION       & SkyScript      & 0.08      & 0.32      & 0.62      & 0.18      & 0.58      & 1.24   & 0.50  \\
		SkyCLIP      & ViT-L-14       & LAION       & SkyScript       & 0.08      & 0.36      & 0.74      & 0.20      & 0.90      & 1.40   & 0.61  \\
		RemoteCLIP   & ResNet-50      & LAION    & R3+D10+S4    & 0.08      & 0.28      & 0.52      & 0.24      & 0.48      & 0.88  & 0.41    \\
		RemoteCLIP   & ViT-B-32       & LAION    & R3+D10+S4    & 0.02      & 0.22      & 0.44      & 0.10      & 0.32      & 0.58  & 0.28    \\
		RemoteCLIP   & ViT-L-14       & LAION    & R3+D10+S4    & 0.04      & 0.38      & 0.64      & 0.10      & 0.62      & 1.00  & 0.46    \\ \hline
		SARCLIP-ISPRS & ResNet-50  & LAION      & SARCAP          & 0.04      & 0.30     & 0.70    & 0.24   & 0.60     & 1.18   & 0.51 \\
		SARCLIP      & ResNet-50      & LAION         & SARVLM-1M & 8.52      & 26.04     & 36.90     & 9.38   & 28.38     & 39.28   & 24.75 \\		
		SARCLIP $^{\ddagger}$ & ResNet-50 & R3+D10+S4 & SARVLM-1M & 8.56  & 25.50     & 37.66    & 9.22   & 28.00    & 39.78  & 24.79 {\scriptsize{(\textcolor{MyForestGreen}{+0.04}}}) \\ \hline
		SARCLIP-ISPRS        & ViT-B-32  & --      & --           & 0.06      & 0.82     & 1.54    & 0.26   & 1.00     & 1.76   & 0.91 \\
		SARCLIP            & ViT-B-32  & LAION      & SARVLM-1M   & 10.20      & 28.54     & 41.26    & 10.84   & 29.86     & 42.18   & 27.15 \\
		SARCLIP $^{\dagger}$ & ViT-B-32  & RS5M    & SARVLM-1M    & 10.80  & 28.86     & 41.14    & 11.30   & 31.08    & 42.78   & 27.66 {\scriptsize{(\textcolor{MyForestGreen}{+0.51})}} \\
		SARCLIP $^{\S}$ & ViT-B-32  & SkyScript    & SARVLM-1M    & 9.70  & 28.72     & 40.16    & 10.56   & 30.32    & 41.88   & 26.89 {\scriptsize{(-0.26)}} \\
		SARCLIP $^{\ddagger}$ & ViT-B-32  & R3+D10+S4 & SARVLM-1M & 10.44  & 29.70     & 41.04    & 10.16   & 30.96    & 42.52   & 27.47 {\scriptsize{(\textcolor{MyForestGreen}{+0.32})}} \\ \hline
		SAR-TEXT      & ViT-L-14       & LAION    & SARTEXT    & 0.48      & 1.82     & 2.90   & 0.42      & 1.38     & 2.12   & 1.52 \\
		SARCLIP-ISPRS & ViT-L-14       & LAION    & SARCAP    & 0.30      & 1.36    & 2.40   & 0.46      & 1.62     & 2.64   & 1.46  \\
		SARCLIP       & ViT-L-14       & LAION    & SARVLM-1M & 11.84      & 32.60     & 44.04   & 12.88      & 33.72     & 44.68  & 29.96 \\
		SARCLIP $^{\dagger}$ & ViT-L-14  & RS5M         & SARVLM-1M   & 12.64      & 33.24     & 44.90   & 13.58      & 34.50     & 46.80  & \textbf{30.94}  {\scriptsize{(\textcolor{MyForestGreen}{+0.98})}}  \\
		SARCLIP $^{\S}$ & ViT-L-14  & SkyScript    & SARVLM-1M    & 12.36  & 32.48  & 44.10  & 12.86  & 34.54  & 45.80   & 30.36 {\scriptsize{(\textcolor{MyForestGreen}{+0.40})}}  \\ 
		SARCLIP $^{\ddagger}$ & ViT-L-14  & R3+D10+S4 & SARVLM-1M & 12.66  & 32.98  & 44.14  & 12.88  & 34.60  & 45.46    & \underline{30.45} {\scriptsize{(\textcolor{MyForestGreen}{+0.49})}} \\ \hline
	\end{tabular*}
	\label{result-retrieval}
\end{table*}
\section{Experiment}

\subsection{Datasets}
In this section, the datasets used in this work are introduced. They are grouped into two categories, i.e., pretraining datasets and downstream datasets.

For pretraining, LAION~\cite{OpenCLIP} is used as a natural image--text dataset. RS5M~\cite{RS5M}, SkyScript~\cite{Skyscript}, R3+D10+S4~\cite{RemoteCLIP}, and HQRS-210K~\cite{HQRS-210K} are optical remote sensing image--text datasets derived from GeoRSCLIP, SkyCLIP, RemoteCLIP, and HQRS-CLIP, respectively. In the SAR domain, SARTEXT~\cite{SAR-TEXT} and SARCLIP~\cite{SARCLIP-ISPRS} are existing image--text datasets, while the proposed SARVLM-1M provides a large-scale SAR image--text dataset.

The downstream datasets are divided into six groups: 1) image--text retrieval, 2) linear probing classification, 3) zero-shot classification, 4) semantic localization, 5) SAR image captioning, and 6) object detection. The datasets are described as follows.

\textbf{1) Image--text retrieval: SARVLM-1M test.} 
This SAR retrieval dataset contains 20,628 image--text pairs collected from existing datasets. The test split includes 180 images from MSTAR, 6,667 images from ATRNet-STAR, 2,783 images from SARDet-100K, 7,096 images from FAIR-CSAR, and 3,902 images from SARLANG-1M-Captions. In our experiments, 5,000 image--text pairs are selected for evaluation, where each image is paired with a unique caption.

\textbf{2) Linear probing classification: MSTAR-SOC, FUSAR-Ship, and SAR-VSA.} 
\textbf{MSTAR-SOC} is a SAR target recognition dataset acquired by an X-band radar in HH polarization mode with a spatial resolution of 0.3 m. It contains 10 military vehicle classes and is divided into four experimental settings following~\cite{AConvNet}. Under the SOC setting, 2,747 training images collected from 17 depression angles and 2,425 test images collected from 15 depression angles are used, with all 10 classes appearing in both sets. 
\textbf{FUSAR-Ship} is a ship recognition dataset containing 15 major ship classes, 98 subclasses, and various non-ship maritime targets~\cite{SARATR}. It is constructed from 126 Gaofen-3 images captured in ultrafine-resolution mode (1.124 $\times$ 1.728 m) with dual polarization (DH and DV), covering open sea, coastal, river, island, and land-background scenes. 
\textbf{SAR-VSA} is a fine-grained SAR target recognition dataset with 25 categories, constructed from MSTAR~\cite{MSTAR}, FUSAR-ship~\cite{FUSAR-ship}, and SAR-ACD~\cite{SAR-ACD}. It contains 11,045 training images and 8,161 test images~\cite{SARATR}.

\textbf{3) Zero-shot classification: RESISC45, UC-Merced, and PatternNet.} 
\textbf{RESISC45}~\cite{RESISC45}, \textbf{UC-Merced}~\cite{UC-Merced}, and \textbf{PatternNet}~\cite{PatterNet} are used to evaluate the effectiveness of the proposed two-stage training strategy and its zero-shot generalization capability.

\textbf{4) Semantic localization: AIR-SLT.} 
\textbf{AIR-SLT} is a remote sensing semantic localization benchmark containing 22 large-scale images and 59 text queries, each associated with ground-truth bounding boxes~\cite{SeLo, SeLov2}.

\textbf{5) SAR image captioning: SARVLM-1M test.} 
The SARVLM-1M test split is also used to evaluate the captioning model, which contains 5,000 image--text pairs.

\textbf{6) Object detection: SARDet-100K, SAR-Aircraft, SSDD, and RSAR.} 
\textbf{SARDet-100K}~\cite{SARDet-100K}, \textbf{SAR-Aircraft}~\cite{SA-Net}, and \textbf{SSDD}~\cite{SSDD} are used for downstream object detection, while \textbf{RSAR}~\cite{RSAR} is used for oriented object detection.

\subsection{Evaluation Metrics}
We evaluate the proposed framework on image-text retrieval and target recognition tasks. For image-text retrieval, Recall at top-$K$ ($\mathrm{R@K}$) is adopted as the evaluation metric, where $K \in \{1,5,10\}$. It is defined as
\begin{equation}
	\mathrm{R@K} = \frac{1}{N} \sum_{i=1}^{N} I\left[y_i \in \mathrm{TopK}(q_i)\right],
\end{equation}
where $N$ denotes the number of queries, $y_i$ denotes the ground-truth item corresponding to query $q_i$, and $I[\cdot]$ denotes the indicator function. To provide a more comprehensive evaluation of bidirectional retrieval performance, Mean Recall ($\mathrm{mR}$) is further adopted, which is defined as the average of $\mathrm{R@1}$, $\mathrm{R@5}$, and $\mathrm{R@10}$ over both text-to-image (T2I) and image-to-text (I2T) retrieval:
\begin{equation}
	\mathrm{MeanRecall} = \frac{1}{6} \sum_{m \in \{\mathrm{T2I}, \mathrm{I2T}\}} \sum_{K \in \{1,5,10\}} \mathrm{R@K}_{m}.
\end{equation}
For the target recognition task, accuracy ($\mathrm{ACC}$) is used as the evaluation metric, which is defined as
\begin{equation}
	\mathrm{ACC} = \frac{1}{N} \sum_{i=1}^{N} I\left[\hat{y}_i = y_i\right],
\end{equation}
where $\hat{y}_i$ denotes the predicted label and $y_i$ denotes the ground-truth label. For the semantic localization task, we adopt Rsu, Rda, Ras, and Rmi as evaluation metrics, following~\cite{SeLo}. For the image captioning task, we employ BLEU~\cite{bleu}, ROUGE\_L~\cite{rouge_L}, METEOR~\cite{meteor}, CIDEr~\cite{cider}, and BERTScore~\cite{BERTScore} as evaluation metric. For object detection task, mAP, mAP$_{50}$, and mAP$_{75}$ are utilized as evaluation metric.
\begin{table}[t!]
	\setlength{\tabcolsep}{0.5mm}
	\renewcommand{\arraystretch}{1.1} 
	\centering
	\caption{Recognition results on MSTAR-SOC and SAR-VSA dataset and improvements over the SARCLIP-ISPRS$(\mathbf{\%})$.}
	\begin{tabular*}{\hsize}{@{\extracolsep{\fill}}llccc}
		\hline
		\multirow{2}{*}{Method} & \multirow{2}{*}{Backbone}   & \multirow{2}{*}{Param.} & \multicolumn{2}{c}{dataset} \\  
		\cmidrule(lr){4-5}
		&    &   & SOC     & VSA       \\  \hline
		OpenCLIP          & ViT-L-14         & 304    & 79.54        & 81.75       \\
		RemoteCLIP        & ViT-L-14         & 304     & 76.74       & 80.01       \\
		SAR-TEXT		  & ViT-L-14         & 304     & 74.55       & 75.15       \\
		SARCLIP-ISPRS     & ViT-L-14         & 304     & 54.47       & 67.71       \\
		SARCLIP $^{\dagger}$  & ViT-L-14     & 304     & \underline{84.74}       & 86.25       \\
		SARCLIP $^{\S}$       & ViT-L-14     & 304     & 82.92       & \textbf{87.57}          \\ 
		SARCLIP $^{\ddagger}$ & ViT-L-14     & 304     & \textbf{86.55}   & \underline{87.29}  \\
		\multicolumn{3}{l}{$\Delta$ (SARCLIP $^{\ddagger}$ vs. SARCLIP-ISPRS)}     & \textcolor{MyForestGreen}{+32.08}  & \textcolor{MyForestGreen}{+19.58}     \\ \hline
	\end{tabular*}
	\label{result-linear-prob}
\end{table}
\subsection{Implementation Details}
We develop SARCLIP based on the OpenCLIP framework. Automatic mixed-precision (AMP) training is employed to reduce memory usage. We adopt ResNet-50, ViT-B-32, and ViT-L-14 as image backbones, with learning rates set to 5e-4, 5e-5, 5e-5, and 5e-5, respectively. ResNet-50 is trained for 30 epochs, while ViT-B-32, ViT-B-16, and ViT-L-14 are trained for 10 epochs. The batch size is set to 256. Training is accelerated using the Adam optimizer, combined with a linear warm-up and cosine learning rate schedule. Downstream experiments are conducted using the wise-ft \cite{wise-ft} and SLM\cite{SeLo} framework. For the downstream target recognition task, we freeze the image backbone and fine-tune only the linear classification layers for 10,000 epochs. All experiments are conducted on two 80GB NVIDIA H100 GPUs. For SAR imagery caption task, we employ optical image caption dataset HQRS-210K as pretrained data.

For object detection, the visual encoder(ViT-B-16) of SARVLM is used as the backbone with a Faster R-CNN head. For oriented object detection, the same visual encoder is combined with an oriented head. All detection experiments are implemented based on MMDetection and MMRotate.
\begin{table}[t!]
	\setlength{\tabcolsep}{1mm}
	\renewcommand\arraystretch{1.1}
	\centering
	\caption{Zero-shot classification results on SAR target recognition dataset and improvements over SARCLIP-ISPRS $(\mathbf{\%})$.}
	\begin{tabular*}{\hsize}{@{\extracolsep{\fill}}llcccc}
		\hline  
		\multirow{2}{*}{Method} & \multirow{2}{*}{Backbone}   & \multirow{2}{*}{Param.}   & \multicolumn{3}{c}{dataset} \\  
		\cmidrule(lr){4-6}
		&        &        & SOC     & VSA  & FUSAR     \\  \hline
		OpenCLIP          & ViT-L-14       & 304       & 48.54  & 19.53 & 57.93   \\ 
		RemoteCLIP        & ViT-L-14       & 304       & 60.12  & 18.58 & 40.37    \\ 
		SAR-TEXT          & ViT-L-14       & 304       & 51.71  & 35.62 & 74.52   \\
		SARCLIP-ISPRS     & ViT-L-14       & 304       & 53.24  & 30.80 & 66.56   \\
		SARCLIP $^{\dagger}$  & ViT-L-14   & 304       & 53.32  & 36.07 & 80.20   \\
		SARCLIP $^{\S}$       & ViT-L-14   & 304       & 52.29     & 35.43 & 81.05   \\
		SARCLIP $^{\ddagger}$ & ViT-L-14   & 304       & \textbf{60.45}  & \textbf{40.7}9 & \textbf{86.12}    \\ 
		\multicolumn{3}{l}{$\Delta$ (SARCLIP $^{\ddagger}$ vs. SARCLIP-ISPRS)}     & \textcolor{MyForestGreen}{+7.2}  & \textcolor{MyForestGreen}{+9.99} & \textcolor{MyForestGreen}{+19.56}    \\ \hline
	\end{tabular*}
	\label{result-zero-shot-SAR}
\end{table}
\subsection{Experiment Results}
In this section, we compare out method with other state-of-the-art CLP-based method include OpenCLIP \cite{OpenCLIP}, RemoteCLIP \cite{RemoteCLIP}, GeoRSCLIP \cite{RS5M}, SkyCLIP \cite{Skyscript}, SAR-TEXT~\cite{SAR-TEXT}, and SARCLIP-ISPRS~\cite{SARCLIP-ISPRS}.
\begin{table}[t]
	\centering
	\renewcommand{\arraystretch}{1.1} 
	\caption{Zero-shot classification on optical remote sensing image and improvements over SARCLIP-ISPRS $(\mathbf{\%})$.}
	\label{result_zero_shot_optical}
	\setlength{\tabcolsep}{0.5pt}
	\begin{tabular*}{\hsize}{@{\extracolsep{\fill}}lcccccc}
		\hline
		\multirow{2}{*}{Method} & \multicolumn{2}{c}{RESISC45} &  \multicolumn{2}{c}{UC-Merced}  &  \multicolumn{2}{c}{PatternNet} \\ 
		\cmidrule(lr){2-3} \cmidrule(lr){4-5} \cmidrule(lr){6-7}
		& Top1 & Top5 & Top1 & Top5 & Top1 & Top5 \\ \midrule
		SARCLIP-ISPRS                    & 7.29 & 21.38 & 11.09 & 37.81  & 9.09 & 25.28 \\
		SARCLIP (single stage)           & 7.93 & 32.62 & 20.29 & 55.62  & 9.97 & 33.21 \\
		SARCLIP $^{\dagger}$            & 7.71 & 30.27 & 21.14 & 44.05 & 12.33 & \textbf{34.76}   \\ 
		SARCLIP $^{\S}$                  & 8.87 & \textbf{34.20} & 20.14 & \textbf{57.05} & 10.72 & 32.87 \\ 
		SARCLIP${\ddagger}$ (two stage)  & \textbf{9.40} & 33.96 & \textbf{23.00} & 55.95 & \textbf{12.98} & 34.07 \\
		vs. SARCLIP-ISPRS & \textcolor{MyForestGreen}{+2.11} & \textcolor{MyForestGreen}{+12.58} & \textcolor{MyForestGreen}{+11.91} & \textcolor{MyForestGreen}{+18.14}  & \textcolor{MyForestGreen}{+3.89} & \textcolor{MyForestGreen}{+8.79} \\
		\hline
	\end{tabular*}
\end{table}
\subsubsection{Retrieval Results on SARVLM-1M Dataset} 
As presented in Table~\ref{result-retrieval}, we compare our SARCLIP methods with several state-of-the-art visual-language foundation models (VLFMs) designed for natural scene and remote sensing images. The results show that models pretrained solely on natural scene data (e.g., LAION) or optical remote sensing data (e.g., RS5M, SkyScript, R3+D10+S4) perform poorly on SAR image-text retrieval. For example, OpenCLIP variants achieve mean recalls below 0.49\%, while GeoRSCLIP, SkyCLIP and RemoteCLIP variants remain below 0.61\%, demonstrating that direct transfer from natural or optical domains is largely ineffective for SAR tasks. In contrast, our two-stage training strategy, which first leverages knowledge from optical remote sensing datasets and then fine-tunes on the SARVLM-1M dataset, substantially improves retrieval performance. Specifically, SARCLIP$^{\ddagger}$ with a ViT-L-14 backbone achieves 30.45\% mean recall, compared with only 1.46\% for SARCLIP-ISPRS, highlighting a dramatic improvement. Similarly, other SARCLIP variants show consistent gains over baselines, confirming the effectiveness of the proposed domain adaptation and multi-modal alignment strategy for the SAR domain.

\subsubsection{Target Recognition Results on MSTAR-SOC and SAR-VSA Dataset} Table~\ref{result-linear-prob} presents the performance of our SARCLIP models on downstream target recognition tasks after fine-tuning a linear layer, thereby demonstrating their visual understanding capabilities. The experimental results show that our SARCLIP series models consistently achieve superior performance. On the MSTAR-SOC\textcolor{red}{\footnote{To prevent data leakage during linear probing experiments, we excluded data and labels associated with downstream tasks from the pretraining weights. Specifically, when evaluating on the MSTAR-SOC dataset, the MSTAR-SOC, SARSim, and SAMPLE datasets were excluded. Similarly, corresponding datasets were withheld during evaluation on the SAR-VSA dataset.}} dataset, SARCLIP$^{\ddagger}$ achieves an impressive 86.55\% accuracy, significantly outperforming optical VLFMs which typically hover around 80\%. Similarly, on the SAR-VSA dataset, our SARCLIP$^{\dagger}$ reaches 87.29\% accuracy, notably surpassing OpenCLIP's 81.75\%. These results underscore the strong visual feature extraction our SARCLIP models for SAR imagery.
\begin{table}[t]
	\centering
	\footnotesize
	\renewcommand{\arraystretch}{1.1}
	\setlength{\tabcolsep}{0.5pt}
	\caption{Object detection result on three SAR datasets.}
	\label{tab:sar_detection_results}
	\begin{tabular*}{\hsize}{@{\extracolsep{\fill}}lccc}
		\toprule
		\multicolumn{4}{c}{\textbf{SARDet-100K (Object detection)}} \\
		\midrule
		Method & mAP $\uparrow$ & mAP$_{50}$ $\uparrow$ & mAP$_{75}$ $\uparrow$ \\
		\midrule
		DETR~\cite{DETR}                         & 31.8 & 62.3 & 30.0 \\
		Sparse R-CNN~\cite{Sparse-RCNN}          & 38.1 & 62.3 & 38.8 \\
		Dab-DETR~\cite{DABDETR}                  & 45.9 & 79.0 & 47.9 \\
		Deformable DETR~\cite{deformable_detr}   & 50.0 & 85.1 & 51.7 \\
		Swim Transformer~\cite{swin}                         & 53.8 & 87.8 & 59.0 \\
		VAN~\cite{van}                                      & 53.5 & 86.8 & 58.0 \\
		ConvNetX~\cite{convnet}                                 & 55.1 & 87.8 & 59.5 \\
		MSFA~\cite{SARDet-100K}                  & 56.4 & 88.2 & 59.5 \\
		DenoDet v2~\cite{DenoDetv2}              & 56.7 & -- & -- \\
		SARATR-X~\cite{SARATR}                   & \underline{57.3} & \underline{88.7} & \underline{62.8} \\
		SARVLMDet(ours)               & \textbf{57.4} {\scriptsize(\textcolor{MyForestGreen}{+0.1})} & \textbf{89.4} {\scriptsize(\textcolor{MyForestGreen}{+0.7})} & \textbf{63.4} {\scriptsize(\textcolor{MyForestGreen}{+0.6})} \\
		
		\midrule
		\multicolumn{4}{c}{\textbf{SAR-Aircraft (Aircraft detection)}} \\
		\midrule
		Method  & mAP $\uparrow$ & mAP$_{50}$ $\uparrow$ & mAP$_{75}$ $\uparrow$ \\
		\midrule
		Cascade R-CNN~\cite{cascade-rcnn}  & - & 75.7 & 58.9 \\
		RepPoints~\cite{Reppoint}          & - & 72.6 & 53.3 \\
		SKG-Net~\cite{SKG-Net}             & - & 70.7 & 46.4 \\
		SA-Net~\cite{SA-Net}               & - & 77.7 & 62.8 \\
		SARATR-X~\cite{SARATR}             & \textbf{58.7} & \underline{86.1} & \underline{64.7} \\
		SARVLMDet(ours)        & \underline{58.3} {\scriptsize{(-0.4)}} & \textbf{86.2} {\scriptsize(\textcolor{MyForestGreen}{+0.1})} & \textbf{66.8} {\scriptsize(\textcolor{MyForestGreen}{+2.1})} \\
		
		\midrule
		\multicolumn{4}{c}{\textbf{SSDD (Ship detection)}} \\
		\midrule
		Method & mAP $\uparrow$ & AP$_{50}$ $\uparrow$ & AP$_{75}$ $\uparrow$ \\
		\midrule
		RetinaNet~\cite{RetinaNet}         & 51.7 & 86.6 & 54.3 \\
		Faster R-CNN~\cite{fasterrcnn}     & 54.4 & 85.2 & 61.1 \\
		Cascade R-CNN~\cite{cascade-rcnn}  & 55.1 & 88.1 & 60.3 \\
		FCOS~\cite{FCOS}                   & 56.0 & 91.9 & 61.7 \\
		FEPS-Net~\cite{FEPS-Net}           & 59.9 & 96.0 & 67.5 \\
		SARVLMDet(ours)         & \textbf{65.1} {\scriptsize(\textcolor{MyForestGreen}{+5.2})} & \textbf{92.1} {\scriptsize(-3.9)} & \textbf{79.8} {\scriptsize(\textcolor{MyForestGreen}{+11.4})} \\
		\bottomrule
	\end{tabular*}
\end{table}
\begin{table}[t]
	\centering
	\renewcommand{\arraystretch}{1.1} 
	\setlength{\tabcolsep}{0.5pt}
	\caption{Oriented SAR object detection result on RSAR dataset.}
	\label{tab_result_rsar}
	\begin{tabular*}{\hsize}{@{\extracolsep{\fill}}llccc}
		\toprule
		\multicolumn{2}{c}{Method}  & mAP $\uparrow$ & AP$_{50}$ $\uparrow$ & AP$_{75}$ $\uparrow$ \\
		\midrule
		\multirow{2}{*}{HBB-based}
		& H2RBox~\cite{h2rbox}             & 18.29 & 49.92 & 11.09  \\
		& H2RBox-v2~\cite{h2rboxv2}          & 32.64 & 68.33 & 26.17   \\
		\midrule
		\multirow{2}{*}{DETR-based}
		& Deformable DETR~\cite{deformable_detr}    & 19.63 & 46.62 & 13.06  \\
		& ARS-DETR~\cite{arsdetr}           & 31.56 & 61.14 & 28.97  \\
		\midrule
		\multirow{4}{*}{One-stage}
		& RetinaNet~\cite{RetinaNet}  & 27.65 & 57.67 & 22.72 \\
		& R$^3$Det~\cite{r3det}   & 30.50 & 63.94 & 25.02  \\
		& S$^2$ANet~\cite{s2anet}  & 33.11 & 66.47 & 28.52 \\
		& FCOS~\cite{FCOS}       & 34.22 & 66.66 & 31.45  \\
		\midrule
		\multirow{5}{*}{Two-stage}
		& Faster RCNN~\cite{fasterrcnn}  & 30.46 & 63.18 & 24.88  \\
		& Oriented RCNN~\cite{orientedrcnn}       & 33.62 & 64.82 & 32.69 \\
		& ReDet~\cite{redet}        & 34.30 & 64.71 & 32.84 \\
		& RoI Transformer~\cite{roitransformer} & \underline{35.02} & \underline{66.95} & \underline{32.65}  \\
		& SARVLMRot(ours)          & \textbf{36.02} {\scriptsize(\textcolor{MyForestGreen}{+1.00})} & \textbf{69.60} {\scriptsize(\textcolor{MyForestGreen}{+2.65})} & \textbf{33.80} {\scriptsize(\textcolor{MyForestGreen}{+1.15})} \\
		\bottomrule
	\end{tabular*}
\end{table}
\begin{figure}[t!]
	\centering
	\includegraphics[width=0.9\linewidth]{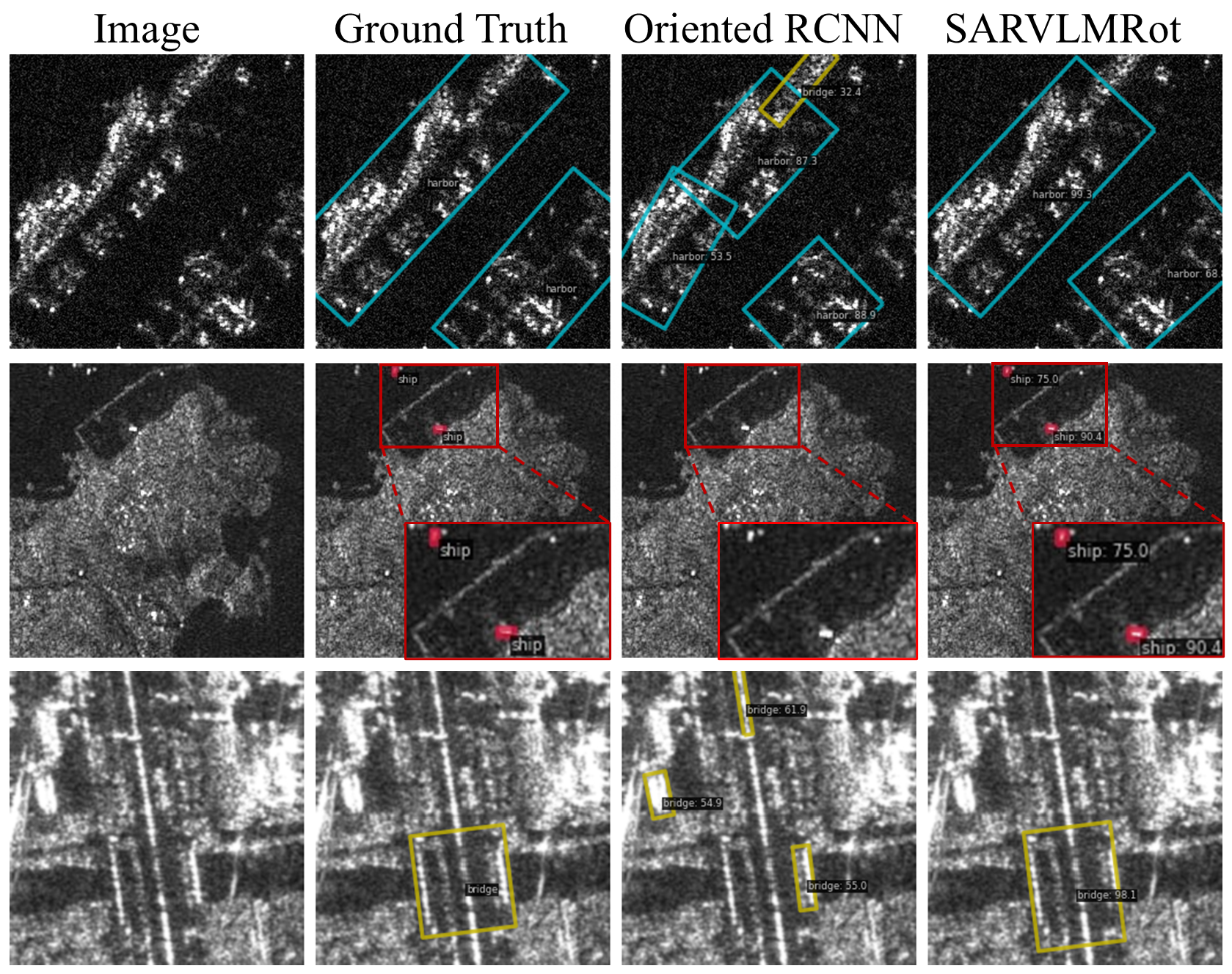}
	\caption{Visualization of oriented object detection on RSAR dataset.}
	\label{fig_rot}
\end{figure}
\subsubsection{Zero-shot Classification on SAR Images} 
Table~\ref{result-zero-shot-SAR} reports the zero-shot classification performance of various models on three SAR target recognition benchmarks: MSTAR-SOC, SAR-VSA, and FUSAR-Ship. Models pretrained on natural or optical domains, such as OpenCLIP and RemoteCLIP, achieve moderate to low accuracies, indicating limited generalization to SAR imagery. SARCLIP-ISPRS improves over these baselines, but the gains remain modest. By contrast, our two-stage SARCLIP$^{\ddagger}$, which combines pretraining on optical remote sensing datasets with fine-tuning on SARVLM-1M, consistently outperforms all competitors across all datasets, achieving Top-5 accuracy at 60.45\%, 40.79\%, and 86.12\% on MSTAR-SOC, SAR-VSA, and FUSAR-Ship\textcolor{red}{\footnote{The same precaution against data leakage was applied during zero-shot experiments.}}, respectively. Compared with SARCLIP-ISPRS, this corresponds to substantial improvements of +7.2\%, +9.99\%, and +19.56\%, demonstrating that the proposed training strategy effectively transfers knowledge to SAR tasks and enhances zero-shot recognition performance.

\subsubsection{Object detection on SAR dataset} The results in Table~\ref{tab:sar_detection_results} demonstrate the effectiveness of SARVLMDet on SAR object detection\textcolor{red}{\footnote{None of the test sets from the downstream detection datasets are included in the pre-training.}}. On SARDet-100K, the proposed method achieves the best performance with 57.4 mAP, 89.4 mAP$_{50}$, and 63.4 mAP$_{75}$, outperforming SARATR-X by 0.1\%, 0.7\%, and 0.6\%, respectively. On SAR-Aircraft, improved mAP$_{50}$ at 86.20\% and mAP$_{75}$ at 66.80\% are obtained. On SSDD, SARVLMDet obtains the best overall results, particularly with a notable gain of 11.4 points in AP$_{75}$. This verifies that the proposed framework can provide strong and transferable priors for downstream SAR object detection.

\subsubsection{Oriented object detection on RSAR}
Table~\ref{tab_result_rsar} demonstrate the effectiveness of the proposed SARVLMRot for oriented SAR object detection\textcolor{red}{\footnote{None of the test sets from RSAR are included in the pre-training.}}. Specifically, SARVLMRot achieves the best performance among all compared methods, reaching 36.02 mAP, 69.60 AP$_{50}$, and 33.80 AP$_{75}$. Compared with RoI Transformer, SARVLMRot improves mAP by 1.00\%, AP$_{50}$ by 2.65\%, and AP$_{75}$ by 1.15\%. It also consistently outperforms representative HBB-based, DETR-based, one-stage, and two-stage detectors. These results indicate that the semantic representations learned by SARVLM provide stronger discrimination and localization capability for oriented targets in complex SAR scenes. Moreover, the detection result are shown in Fig.~\ref{fig_rot}. This again verifies that the proposed framework can serve as an effective prior for downstream oriented SAR object detection. 
\subsubsection{Zero-shot Classification on Optical Remote Sensing Images} 
Table~\ref{result_zero_shot_optical} reports the zero-shot classification performance of our SARCLIP on three optical remote sensing benchmarks: RESISC45, UC-Merced, and PatternNet. Models pretrained solely on SAR data (SARCLIP-ISPRS) achieve relatively low accuracy, with Top-1 scores ranging from 7.29\% to 11.09\% and Top-5 scores from 21.38\% to 37.81\%, indicating limited transferability to optical domains. In contrast, SARCLIP that incorporate two-stage training exhibit clear improvements. SARCLIP$^{\ddagger}$, which leverages pretraining on optical remote sensing datasets followed by fine-tuning on SARVLM-1M, achieves the highest Top-1 accuracies across all three datasets (9.40\%, 23.00\%, and 12.98\%) and competitive Top-5 scores, yielding substantial gains over the SARCLIP-ISPRS baseline (e.g., +12.58\% on RESISC45 Top-5 and +18.14\% on UC-Merced Top-5). These results demonstrate that the two-stage pretraining strategy not only improves SAR image-text alignment but also enhances the generalization ability of the model to optical remote sensing imagery.
\begin{table}[t!]
	\setlength{\tabcolsep}{1mm}
	\renewcommand{\arraystretch}{1.1} 
	\centering
	\caption{Semantic Localization results on AIR-SLT datase.}
	\label{result-senmatic-localization}
	\begin{tabular*}{\hsize}{@{\extracolsep{\fill}}llcccc}
		\hline
		Method            & Backbone      & Rsu $\uparrow$  & Rda $\uparrow$  & Ras $\downarrow$  & Rmi$\uparrow$  \\  \hline 
		RemoteCLIP        & ViT-L-14       & 0.7707  & 0.6722   & 0.2876   & 0.7257      \\ 
		SARCLIP $^{\dagger}$  & ViT-L-14    & 0.5654   & 0.4531   & 0.5601   & 0.4889    \\
		SARCLIP $^{\S}$       & ViT-L-14    & 0.5675   & 0.4792   & 0.5677   & 0.4981    \\
		SARCLIP $^{\ddagger}$ & ViT-L-14    & 0.5741  & 0.4173   & 0.5731   & 0.4834    \\ \hline
	\end{tabular*}
\end{table}
\begin{figure}[t!]
	\centering
	\includegraphics[width=0.48\textwidth]{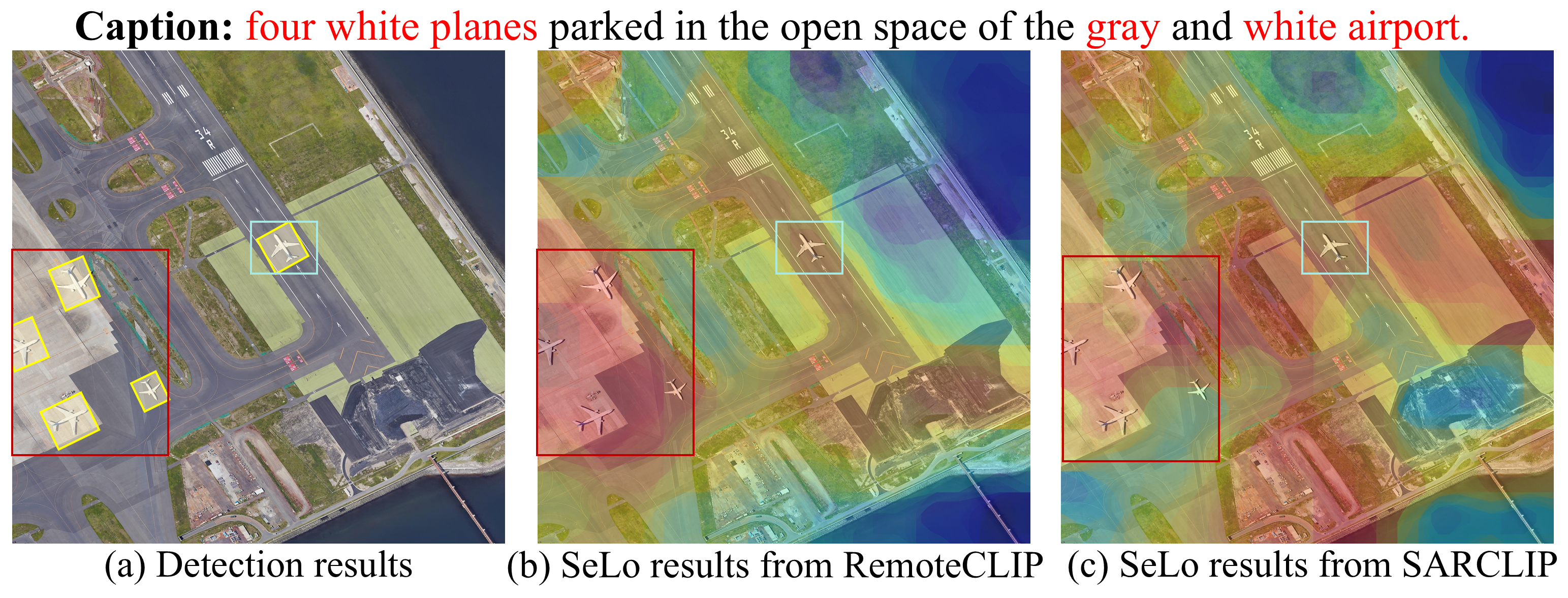}
	\caption{Visualization of results on semantic localization task from SARCLIP (ViT-L-14). The red boxes indicate correct results, and the cyan boxes indicate incorrect results.}
	\label{fig_SeLo_result}
\end{figure}
\subsubsection{Semantic Localization  on AIR-SLT Dataset}
Table~\ref{result-senmatic-localization} reports semantic localization results on the optical AIR-SLT dataset under a cross-domain setting. Our SARCLIP$^{\ddagger}$ achieves 0.5741 Rsu compared to 0.7707 for RemoteCLIP, showing that SARCLIP retains a certain level of generalization to optical scenes. As illustrated in Fig.~\ref{fig_SeLo_result}, given query captions four white planes and gray and white airport, our model successfully captures the plane features (red boxes in Fig.~\ref{fig_SeLo_result}) but fails to identify the gray and white airport region (cyan boxes in Fig.~\ref{fig_SeLo_result}). This limitation arises from the intrinsic differences between SAR and optical imagery, since SAR backbone models cannot perceive color, although they are capable of capturing object textures and shapes.
\begin{table*}[t!]
	\centering
	\renewcommand{\arraystretch}{1.1} 
	\setlength{\tabcolsep}{0.1pt}
	\caption{Experiment result on image caption task.}
	\label{result-caption}
	\begin{tabular*}{\hsize}{@{\extracolsep{\fill}}lccccccccccc}
		\hline
		Method     & Image Encoder    & Pretrain Data & Tune On  & BLUE1    & BLUE2  & BLUE3 & BLUE4 & METEOR & ROUGE\_L & CIDEr  & BERTScore  \\  \hline
		CoCa-Vanilla   & ViT-L-14 & LAION  & --        & 8.75  & 3.21   & 1.14   & 0.32   & 10.54   & 10.67   & 4.17   & 82.23\\ 
		CoCa-HQRS      & ViT-L-14 & LAION  & HQRS      & 10.79 & 4.02   & 1.14   & 0.39   & 8.21    & 13.11   & 7.47   & 84.00 \\
		SARCap         & ViT-L-14 & --  & SARVLM-1M   & 22.42 & 14.44  & 10.45  & 7.79   & 19.23   & 21.76   & 24.32  & 88.11  \\
		SARCap $^{*}$         & ViT-L-14 & LAION+HQRS  & SARVLM-1M  & \textbf{26.10}   & \textbf{17.13}  & \textbf{11.93}   & \textbf{8.47}   & \textbf{22.18}  & \textbf{26.15}   & \textbf{29.58}     & \textbf{88.73}    \\
		\multicolumn{4}{l}{$\Delta$ (SARCap$^{*}$ vs. SARCap)}      & \textcolor{MyForestGreen}{+3.68}   & \textcolor{MyForestGreen}{+2.69}   & \textcolor{MyForestGreen}{+1.48}    & \textcolor{MyForestGreen}{+0.68}    & \textcolor{MyForestGreen}{+2.95}   & \textcolor{MyForestGreen}{+4.39}   & \textcolor{MyForestGreen}{+5.26}   & \textcolor{MyForestGreen}{+0.62}  \\  \hline
	\end{tabular*}
\end{table*}
\subsubsection{SAR imagery captioning on SARVLM-1M} 
\begin{figure}[t!]
	\centering
	\includegraphics[width=0.48\textwidth, keepaspectratio]{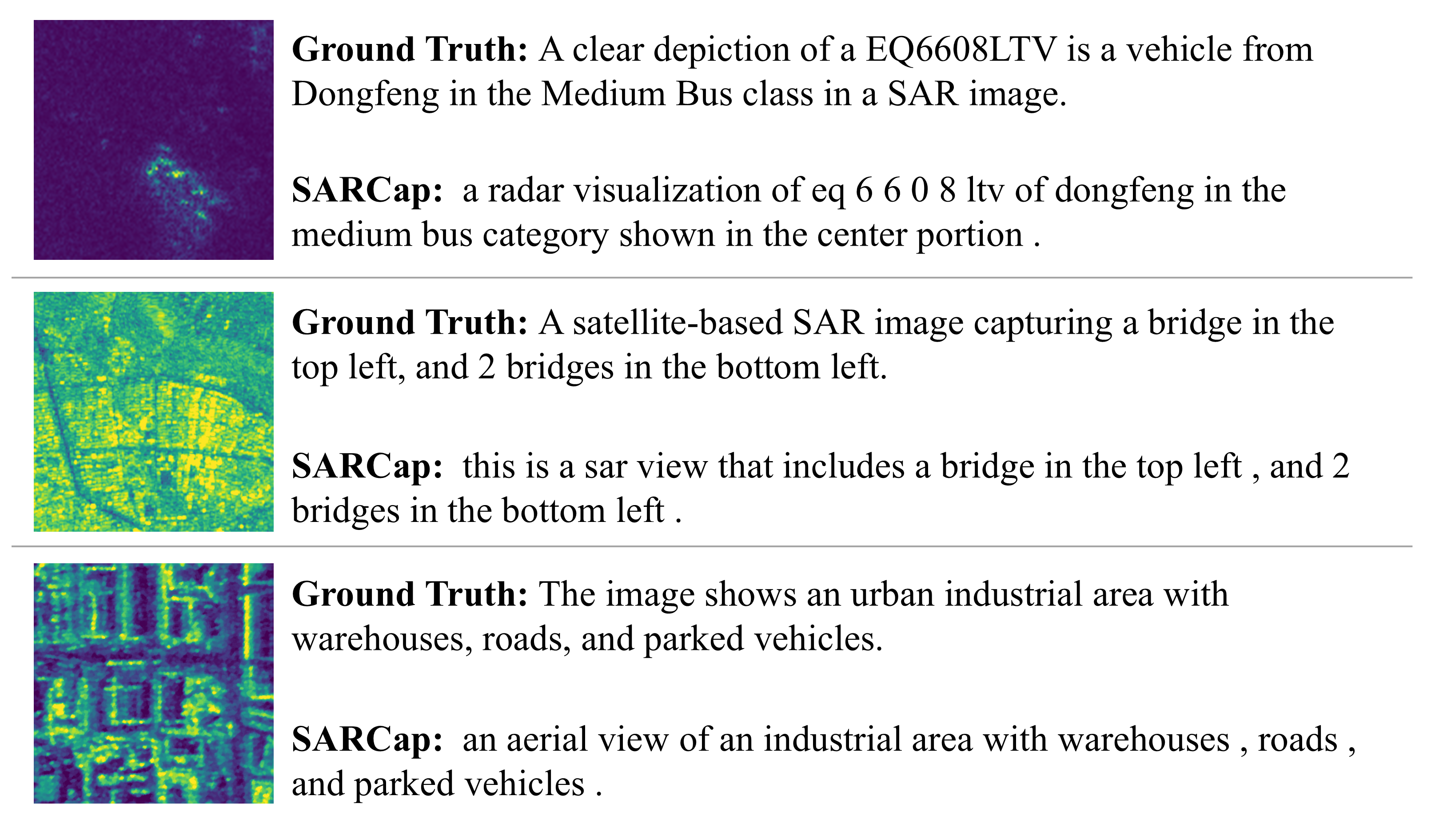}
	\caption{Visualization of SAR imagery caption results from SARCap.}
	\label{fig_caption_result}
\end{figure}

Table~\ref{result-caption} summarizes the captioning performance of our SARCap framework on the SARVLM-1M dataset. Fine-tuning CoCa on high-quality optical remote sensing data (CoCa-HQRS) improves over the baseline CoCa-Vanilla, indicating the benefit of leveraging optical remote sensing domain knowledge. Building on this, we adopt a two-stage progressive training strategy to obtain SARCap$^{*}$, which incorporates both optical pretraining (LAION+HQRS) and SARVLM-1M fine-tuning. SARCap$^{*}$ consistently outperforms the single-stage SARCap across all automatic metrics, including BLEU, METEOR, ROUGE-L, CIDEr, and BERTScore, achieving improvements of up to +5.26\% (CIDEr) and +4.39\% (ROUGE-L). As illustrated in Fig.~\ref{fig_caption_result}, our SARCap model accurately describes vehicles, buildings, and land covers. These results demonstrate that the proposed two-stage strategy effectively transfers cross-domain knowledge and enhances the generation of fluent and semantically meaningful captions for diverse SAR scenes.
\subsection{Ablation Studies}

\subsubsection{Effect of Domain Transferring Strategy} 
Table~\ref{result-ablation-progressive} presents an ablation study on the proposed progressive training strategy. Natural images exhibit a large domain gap with SAR imagery, while optical remote sensing images are closer to SAR, yet transferring knowledge remains challenging. Our two-stage SARCLIP leverages this relationship by pretraining on optical remote sensing datasets and fine-tuning on SARVLM-1M. This strategy leads to notable improvements over the single-stage baseline. Retrieval performance on SARVLM-1M test data increases from 29.96\% to 30.45\%, and zero-shot classification on FUSAR shows the most significant gain, rising from 73.33\% to 86.12\%. Semantic localization on AIR-SLT is also improved, increasing from 56.43\% to 57.41\%. Moreover, our SARCap$^{*}$ model consistently outperforms the single-stage SARCap across all captioning metrics. Notably, it achieves improvements of 5.25 and 0.62 in CIDEr and BERTScore, respectively. These results indicate that the progressive training strategy effectively bridges the domain gap and enhances multi-modal alignment and generalization in the SAR domain.

\subsubsection{Effect of Training Layer} 
\begin{figure}[t!]
	\centering
	\includegraphics[width=\linewidth]{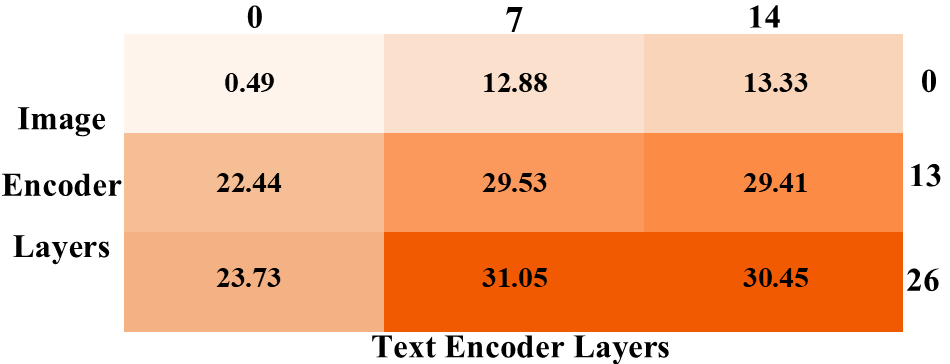}
	\vspace{-3mm}
	\caption{Ablation study on training layers of SARCLIP$^{\ddagger}$ on the SARVLM-1M test set (Mean Recall \%).}
	\label{fig_ablation}
\end{figure}

As shown in Fig.~\ref{fig_ablation}, we evaluate the impact of varying the number of activated layers in the image and text encoders on the SARVLM-1M test set. The results reveal a consistent trend that model performance improves as more layers are activated. This highlights the importance of deep and expressive representations in both modalities for effective SAR image-text understanding.
\subsubsection{Effect of Ensemble Ratio}
\label{Sec_ensemble_ratio}
\begin{figure}[t!]
	\centering
	\includegraphics[width=0.40\textwidth]{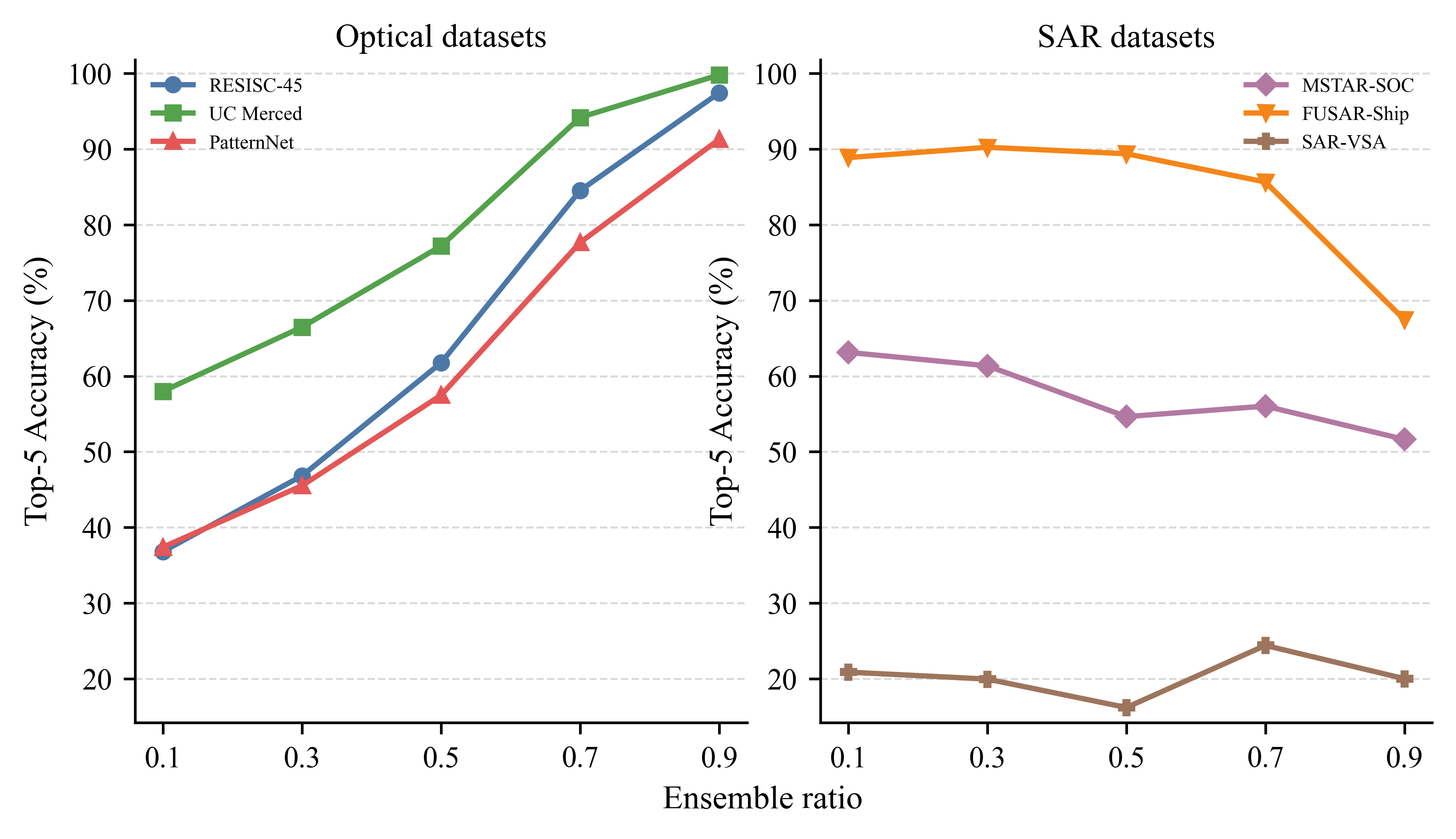}
	\caption{Ablation study on ensemble strategy under different fusion ratios $\alpha$ on representative optical and SAR datasets.}
	\label{fig_ensemble_alpha_curve}
\end{figure}

We perform zero-shot experiments on both optical and SAR datasets with different ensemble ratios to evaluate the effect of $\alpha$. The ratio $\alpha$ is varied from 0.1 to 0.9 with a step size of 0.2. As illustrated in Fig.~\ref{fig_ensemble_alpha_curve}, the Top-5 accuracy on RESISC-45, UC Merced, and PatternNet increases steadily as $\alpha$ becomes larger, whereas the performance on the three SAR datasets generally declines. This opposite trend suggests that the ensemble ratio critically affects the balance between optical and SAR representations. In particular, larger values of $\alpha$ are more beneficial to optical datasets but less suitable for SAR datasets. Therefore, selecting an appropriate $\alpha$ is necessary to obtain a better balance between cross-domain robustness and generalization.

\begin{figure}[t!]
	\centering
	\includegraphics[width=0.45\textwidth]{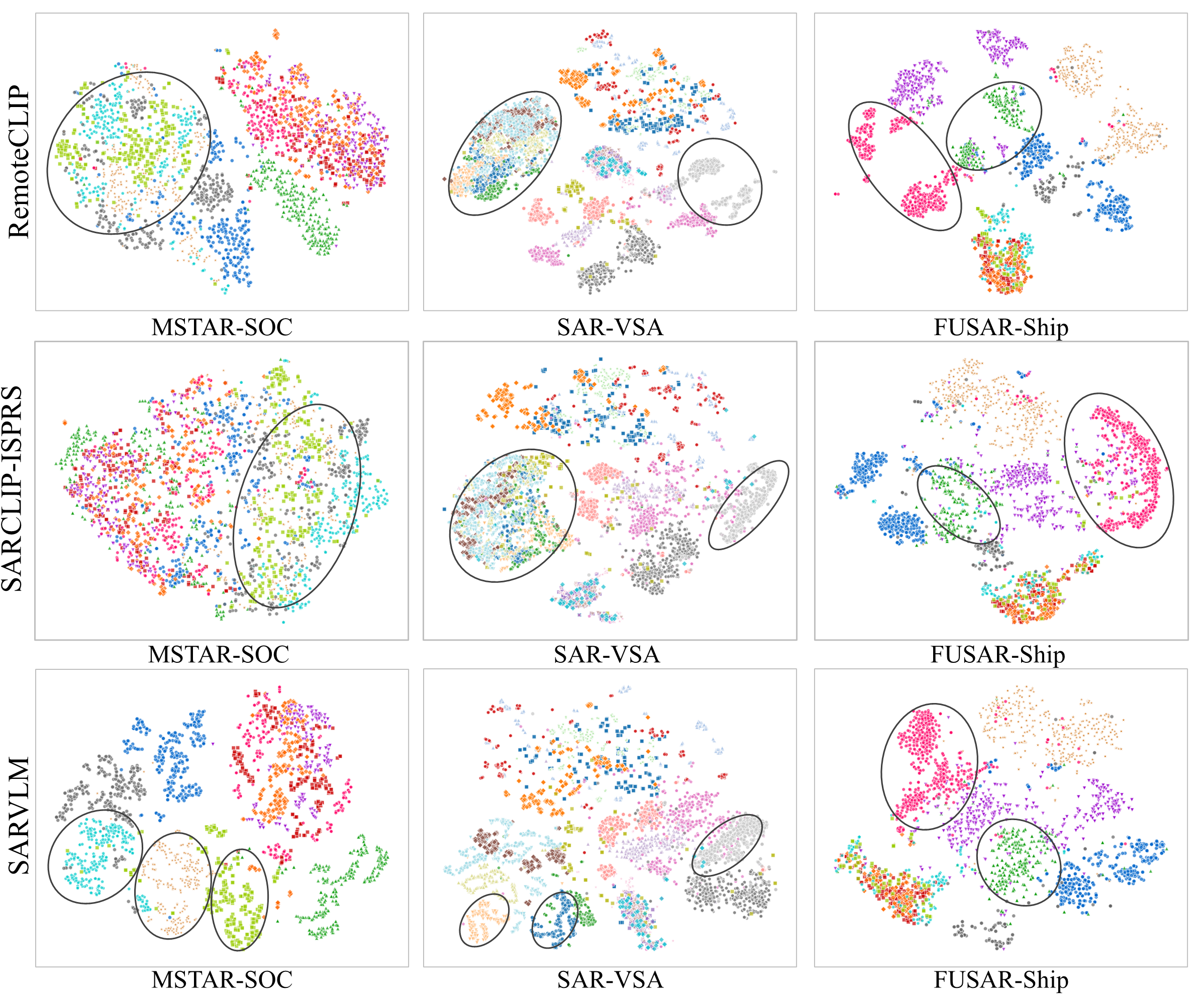}
	\caption{Feature space visualization of RemoteCLIP, SARCLIP-ISPRS and SARCLIP$^{\ddagger}$ image encoder on three downstream datasets (ViT-L-14).}
	\label{fig_visualization}
\end{figure}
\subsection{Visualization}
\subsubsection{Visualization of Attention Heatmap}
\begin{figure*}[t!]
	\centering
	\includegraphics[width=0.95\linewidth]{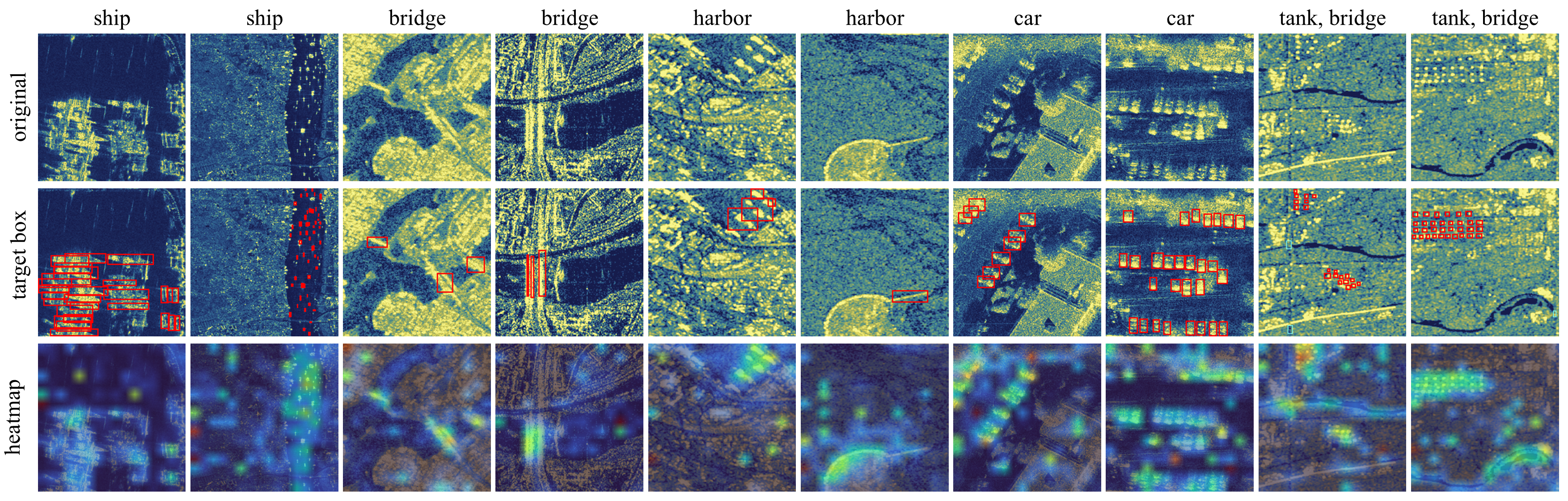}
	\vspace{-2mm}
	\caption{Visualization of the attention maps of the SARVLM image encoder on test images from SARVLM-1M (ViT-L-14).}
	\label{fig_vis_vit}
\end{figure*}

As illustrated in Fig.~\ref{fig_vis_vit}, we present representative SAR scenes selected from the SARVLM validation set\textcolor{red}{\footnote{The selected images do not appear in trainging set.}}, covering five common categories: ship, bridge, harbor, tank, and car. For each sample, the original SAR image, annotated bounding boxes, and the attention heatmap derived from CLS-token attention rollout of the visual transformer are provided. The results show that the model attention is mainly concentrated on the target regions and their surrounding discriminative structures, demonstrating the effectiveness of the learned visual representations for SAR target perception.
\begin{table*}[t!]
	\centering
	\setlength{\tabcolsep}{0.5pt}
	\caption{Ablation study on progressive training strategy with image encoder. RDS represent R3+D10+S4 dataset from RemoteCLIP, while R45, UC, PN, STL denote RSISC-45, UC Merced, PatternNet, AIR-STL dataset, respectively.(ViT-L-14)$(\mathbf{\%})$.}
	\renewcommand\arraystretch{1.1}
	\begin{tabular*}{\hsize}{@{\extracolsep{\fill}}lccccccccccccc}
		\hline
		\multirow{2}{*}{Method}     & \multirow{2}{*}{\makecell{Image\\Encoder}}  & \multirow{2}{*}{\makecell{Pretrain\\Data}}   & \multirow{2}{*}{\makecell{Turn\\ On}}  & Retrieval & \multicolumn{2}{c}{Linear Prob} & \multicolumn{6}{c}{Zero-Shot} & SeLo \\ 
		\cmidrule(lr){5-5} \cmidrule(lr){6-7} \cmidrule(lr){8-13} \cmidrule(lr){14-14}
		&   &  &   & SARVLM-test  & SOC  & VSA & SOC & VSA & FUSAR & R45  & UC  & PN  & STL \\  \hline 
		SARCLIP               & ViT-L-14 & LAION   & SARVLM-1M    & 29.96 & 82.80  & 88.29  & 58.23    & 38.99 & 73.33 & 32.62  & 55.62  & 33.21 & 56.43 \\
		SARCLIP $^{\ddagger}$ & ViT-L-14 & LAION+RDS & SARVLM-1M  & 30.45   & 86.55  & 87.29  & 60.45   & 40.79 & 86.12 & 33.96 & 55.95 & 34.07 & 57.41   \\ 
		\multicolumn{4}{l}{$\Delta$ (SARCLIP $^{\ddagger}$ vs. SARCLIP)}      & \textcolor{MyForestGreen}{+0.49}   & \textcolor{MyForestGreen}{+3.75}   & -1.00    & \textcolor{MyForestGreen}{+2.22}     & \textcolor{MyForestGreen}{+1.8}  & \textcolor{MyForestGreen}{+12.79}  & \textcolor{MyForestGreen}{+1.34}   & \textcolor{MyForestGreen}{+0.33}  & \textcolor{MyForestGreen}{+0.86}  & \textcolor{MyForestGreen}{+0.98} \\  \hline
	\end{tabular*}
	\label{result-ablation-progressive}
\end{table*}
\subsubsection{Visualization of Feature Distribution}
As shown in Fig.~\ref{fig_visualization}, we use t-SNE to visualize RemoteCLIP, SARCLIP-ISPRS, and SARCLIP$^{\ddagger}$ (ViT-L-14) features on the test test of three downstream datasets\textcolor{red}{\footnote{The data used for visualization are excluded from the pretraining set to prevent data leakage and ensure a fair comparison.}}. As illustrated by the ellipses in the figure, the features extracted by our model exhibit a more uniform distribution, with clear boundaries between classes and compact intra-class clustering. For instance, in the feature visualization on the MSTAR, the three classes extracted by RemoteCLIP and SARCLIP-ISPRS exhibit significant overlap, whereas the features from our SARCLIP model are comparatively well-separated. The visualization results indicate that our model provides stronger feature representations for SAR imagery, making it a suitable choice for initialization in downstream tasks.

\vspace{-3mm}
\section{Conclusion}
In this paper, we introduce SARVLM-1M, a large-scale SAR image-text dataset, and propose SARVLM, the first vision-language foundation model tailored for SAR imagery. 
By leveraging a progressive two-stage domain transfer strategy, SARCLIP effectively transfers knowledge from optical remote sensing data to SAR, achieving robust multi-modal alignment, accurate target recognition, zero-shot classification, and semantic localization. SARDet and SARRot demonstrate the effectiveness of the proposed framework for object detection, while SARCap shows strong performance in SAR image captioning. Extensive experiments on multiple datasets validate the effectiveness of our two-stage training approach and highlight the benefits of cross-domain knowledge transfer. 

For future work, we plan to explore the integration of multi-modal large language models with agent-based methods to further enhance SAR image interpretation and reasoning capabilities. 
We anticipate that this line of research will provide a foundation for advanced SAR vision-language applications, including automated analysis, scene understanding, and cross-modal retrieval in remote sensing domains.
\appendices

\ifCLASSOPTIONcaptionsoff
  \newpage
\fi


\bibliographystyle{IEEEtran}
\bibliography{IEEEabrv,refs}

\end{document}